\documentclass[letterpaper]{article} 
\usepackage{aaai25}  
\usepackage{times}  
\usepackage{helvet}  
\usepackage{courier}  
\usepackage[hyphens]{url}  
\usepackage{graphicx} 
\usepackage{natbib}  
\usepackage{caption} 
\usepackage{algorithm}
\usepackage{algorithmic}

\usepackage{newfloat}
\usepackage{listings}

\usepackage{amsmath}
\usepackage{amssymb}
\usepackage{mathtools}
\usepackage{amsthm}
\usepackage{booktabs}
\usepackage{bm}
\usepackage{mathrsfs}
\usepackage{multirow}
\usepackage{tabularx}
\usepackage[integrals]{wasysym}
\usepackage{graphics}
\usepackage{epstopdf}
\usepackage{dutchcal}
\usepackage[T1]{fontenc}
\usepackage{subfigure}
\usepackage{cases}
\newtheorem{theorem}{Theorem}

\theoremstyle{definition}
\newtheorem{definition}{Definition}
\newtheorem{assumption}{Assumption}
\theoremstyle{remark}

\DeclareCaptionStyle{ruled}{labelfont=normalfont,labelsep=colon,strut=off} 
\floatstyle{ruled}
\newfloat{listing}{tb}{lst}{}
\floatname{listing}{Listing}
\title{A Method for Enhancing Generalization of Adam by Multiple Integrations}
\author{
    Long Jin\textsuperscript{\rm 1},
    Han Nong\textsuperscript{\rm 1},
    Liangming Chen\textsuperscript{\rm 1},
    Zhenming Su\textsuperscript{\rm 1}
}
\affiliations{
    \textsuperscript{\rm 1}School of Information Science and Engineering,
    Lanzhou University, Lanzhou, China\\
    jinlongsysu@foxmail.com

}

\usepackage{bibentry}

\begin{document}

 \maketitle
 
\begin{abstract}
The insufficient generalization of adaptive moment estimation (Adam) has hindered its broader application. Recent studies have shown that flat minima in loss landscapes are highly associated with improved generalization.  Inspired by the filtering effect of integration operations on high-frequency signals, we propose multiple integral Adam (MIAdam), a novel optimizer that integrates a multiple integral term into Adam. This multiple integral term effectively filters out sharp minima encountered during optimization, guiding the optimizer towards flatter regions and thereby enhancing generalization capability. We provide a theoretical explanation for the improvement in generalization through the diffusion theory framework and analyze the impact of the multiple integral term on the optimizer's convergence. Experimental results demonstrate that MIAdam not only enhances generalization and robustness against label noise but also maintains the rapid convergence characteristic of Adam, outperforming Adam and its variants in state-of-the-art benchmarks. 
\end{abstract}

%
\begin{links}
    \link{Code}{https://github.com/LongJin-lab/MIAdam}
\end{links}

\section{Introduction}

An appropriate optimizer is essential to train a deep neural network (DNN), as it directly affects the training convergence and performance of a model \cite{yao2021adahessian}. The goal of optimizers is usually to minimize (or maximize) a certain objective function, typically a loss function, which measures the gap between the predictions and ground-truth values. As a traditional optimizer, stochastic gradient descent (SGD)  is a commonly used optimizer for training DNNs \cite{deng2023stability}. However, SGD suffers from certain limitations, such as the need to precisely tune the learning rate, the uniform scaling of gradients in all directions, and the risk of being trapped in saddle points \cite{johnson2020adascale,ziyin2021sgd}. In order to address these challenges, adaptive learning rate optimizers are developed, offering more nuanced control over learning rates and improved convergence in diverse training scenarios. Among them, adaptive moment estimation (Adam) \cite{kingma2014adam} is currently one of the most popular adaptive learning optimizers for its rapid convergence and efficient handling of sparse gradients. The combination of first-order and second-order moments in Adam enables the effective incorporation of momentum-based optimization and adaptive learning rate methods, thereby enhancing its overall efficiency and applicability in various neural network training contexts. 
 \begin{figure}[t]
    \centering
        \includegraphics[width=1\linewidth]{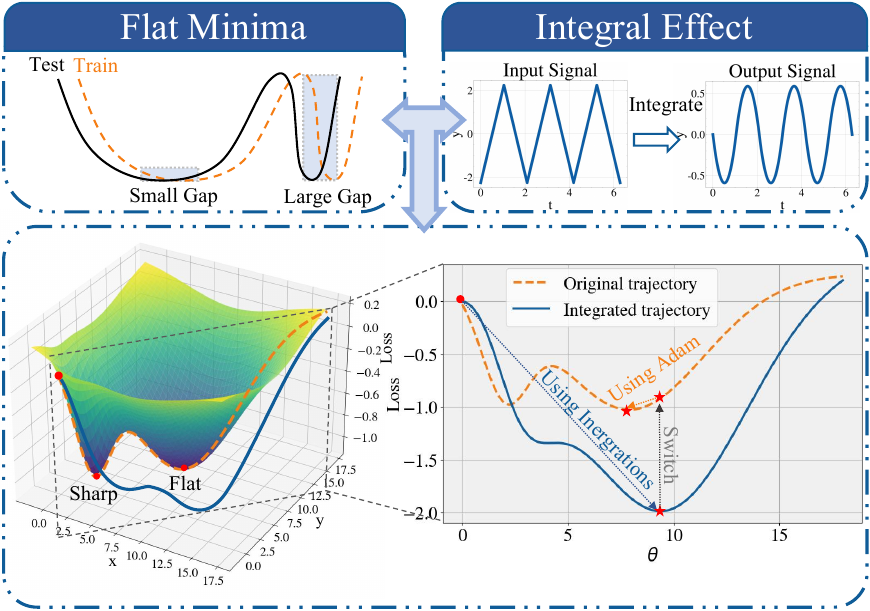} 
      \caption{The idea of this work and the filtering effect of integrations on optimizer trajectories. The blue integrated trajectory represents an equivalent path that does not actually exist on the original loss landscape.}
          \label{fig.1}
  \end{figure}
 Despite being widely used, Adam also exhibits certain limitations, such as the inferior generalization capabilities compared to SGD in some scenarios \cite{wilson2017marginal,luo2019adaptive,zou2021understanding}. Therefore, several enhanced variants of Adam are developed to alleviate the issue of poor generalization. Switching from Adam to SGD (SWATS) \cite{keskar2017improving} is designed to start training with the Adam optimizer, then automatically switch to SGD, aiming to improve the model's generalization performance. However, this method can not maintain the original convergence rate of Adam to some extent. ND-Adam (normalized direction-preserving Adam) \cite{zhang2018improved} meticulously preserves the direction of gradients for each parameter and produces the regularization effect akin to L2 weight decay. Despite this, its ability to enhance generalization is quite limited. AdaBound  \cite{luo2019adaptive} employs dynamic constraints on the learning rate to achieve a smooth and gradual transition from adaptive methods to SGD, which enhances the generalization of a model and reduces the dependence on detailed learning rate adjustments. Nonetheless, a significant drawback of AdaBound is its potential for slow convergence in certain scenarios \cite{savarese2019convergence}. Overall, these improved optimizers of Adam attempting to enhance the generalization of Adam are unable to simultaneously retain the rapid convergence characteristic of Adam and enhance generalization effectively.
 
 Many studies show theoretically and empirically that the generalization performance of a model is highly correlated with its loss landscape in the parameter space \cite{hochreiter1997flat,chaudhari2019entropy,Jiang2020Fantastic,petzka2021relative, du2022sharpness}. A crucial observation is that flat regions in the loss landscape tend to be associated with good generalization performance, while sharp or narrow regions may lead to overfitting \cite{mulayoff2020unique,sun2023dynamic}. This means that optimizers can effectively improve the generalization of a model by converging to flat minima during the training process, which provides a new perspective to alleviate the issue of the poor generalization of Adam. In order to improve the generalization of a model by finding flat minima in the loss landscape, we propose multiple integral Adam (MIAdam), which is inspired by the effect that the multiple integral term can often serve as filters and noise suppressions in the field of signal processing and control systems \cite{roberts1987digital,jin2015integration}. MIAdam introduces a multiple integral term to the parameter update formula of Adam, utilizing the filtering effect of multiple integrations to smooth the optimizer's trajectory. As shown in Fig. \ref{fig.1}, if we consider the trajectory as a time-varying input signal, integrating the signal is equivalent to filtering out the sharp minima encountered by the optimizer on the loss landscape, thereby enabling the optimizer to converge to flat minima. More details about the design of the MIAdam optimizer are discussed in Section MIAdam. The main contributions of this paper are summarized as follows.
\begin{itemize}
\item[$\bullet$] To the best of our knowledge, the method of introducing multiple integral term in the optimizer to find flat minima in the loss landscape is proposed for the first time. Furthermore, we propose a new optimizer based on Adam, which is called MIAdam.
\item[$\bullet$] We provide theoretical analyses on MIAdam. Specifically, utilizing the diffusion theory framework in \cite{xie2020diffusion}, 
we prove that the multiple integral term enables MIAdam to generalize better than Adam under some assumptions. 
In addition, we also analyze the effect of multiple integral terms on convergence. 
\item[$\bullet$] The effectiveness of the proposed method is validated through image classification experiments, text classification experiments, and experiments that inject label noises into datasets. Experimental results demonstrate that MIAdam outperforms the Adam and its state-of-the-art (SOTA) variants on both generalization and robustness against label noises. 
\end{itemize}

\section{Preliminaries}
In this section, core concepts about Adam and the related theoretical analyses on the relationship between flat minima and generalization are briefly given to set the stage for the detailed exposition of MIAdam that follows.
\subsection{Overview of Adam}
The training procedure for a DNN can be primarily characterized as an optimization problem, which is defined as follows:
\begin{equation}
\label{eq1}
\min_{\bm{\theta}}\frac{1}{\left|S\right|}\sum_{k=1}^{\left|S\right|}\mathscr{L}(\bm{x}_k, \bm{y}_k;\bm{\theta}),
\end{equation}
where $\mathscr{L}(\bm{x}_k,\bm{y}_k;\bm{\theta})$ represents the loss function; $\bm{\theta}$ denotes the parameter of the model; $\bm{x}_k$ and $\bm{y}_k$ are the input and its corresponding ground-truth label, respectively;  $S$ is a subset of $D$ and $D$ is the training dataset. 
In the early stages of deep learning development, SGD emerges as a prevailing optimizer, with its parameter update formula expressed as follows:
\begin{equation}
\label{eq2}
\theta_{t+1,i}=\theta_{t,i}- \alpha g_{t,i},
\end{equation}
where $\alpha$ represents the learning rate; 
$\theta_{t,i}$ represents the $i$-th dimension of the parameter at discrete time $t$;
$g_{t,i}$ is the gradient with respect to the parameter $\theta_{t,i}$. The gradient is formally defined as  
\begin{equation}
\label{eq3}
g_{t,i}=\frac{\partial L(\theta_{t,i})}{\partial\theta_{t,i}},
\end{equation}
where $L(\bm{\theta}) =(1/\left|S\right|) \sum\nolimits_{k=1}^{\left|S\right|} \mathscr{L}(\bm{x}_k, \bm{y}_k;\bm{\theta})$.
Momentum is a strategy used to expedite convergence of SGD towards minima and escape saddle points on the loss landscape \cite{qian1999momentum}. It is computed by accumulating previous gradients into the current gradient. The parameter update formula of SGD with momentum (SGDM) is shown as
\begin{subnumcases}{}
m_{t,i}=\beta m_{t-1,i}+g_{t,i} \label{eq4a}, \\
\theta_{t+1,i}=\theta_{t,i}- \alpha m_{t,i}, \label{eq4b}
\end{subnumcases} where $m_{t,i}$ denotes the momentum at $t$ and $\beta$ is the hyperparameter is used to trade off between the current gradient and the accumulation of historical gradients. Adam refines the momentum formulation in Eq. (\ref{eq4a}) and introduces an adaptive learning rate achieved through the computation of the first-order and the second-order moments concerning current gradients.  The first-order and second-order moments are calculated by the following expressions:
\begin{subnumcases}{}
m_{t,i}=\beta_1 m_{t-1,i}+(1-\beta_1)g_{t,i}\label{eq5a}, \\       
v_{t,i}=\beta_2 v_{t-1,i}+(1-\beta_2)g_{t,i}^2,\label{eq5b}
\end{subnumcases}
where $v_{t,i}$ denotes the second-order moment at $t$; the  $\beta_1$ and $\beta_2$ are the exponential decay rates used to adjust the first-order and second-order moments, respectively.
Furthermore, the parameter update formula of Adam is expressed as
\begin{equation}
\begin{aligned}
\label{eq7}
\vspace{-16pt}
\theta_{t+1,i}=\theta_{t,i}- \frac{\alpha \hat{m}_{t,i}}{\sqrt{\hat{v}_{t,i}+\epsilon}},
\vspace{-16pt}
\end{aligned}
\end{equation}
where $\hat{m}_{t,i} = m_{t,i} / (1-\beta_1^t)$ and $\hat{v}_{t,i} = v_{t,i} / (1-\beta_2^t)$. The hyperparameter $\epsilon$ is assumed as a small value to prevent division by zero in the denominator. 
\subsection{The Relationship between Flat Minima and Generalization}

The generalization of DNNs has been extensively explored in recent years. In order to understand the phenomenon of generalization of DNNs, some of the existing research delves into understanding the relationship between loss landscapes and the generalization. A correlation between the flatness of the loss landscape and model generalization is revealed in \cite{hochreiter1995simplifying}. Subsequent investigations in \cite{hochreiter1997flat} expand on this correlation and provide a method for identifying flat minima. In~\cite{keskar2016large}, a definition of the sharpness of a specified point on a loss landscape is given. The study in \cite{dinh2017sharp} introduces a reparameterization method and argues that previous sharpness measurements are inadequate for predicting generalization capabilities. Furthermore, it is demonstrated that the generalization capability is influenced by factors such as the batch size, 
higher-order ``smoothness'' terms characterized by the Lipschitz constant of the Hessian matrix, the loss function, and the number of parameters \cite{wang2018identifying}. Based on the above theoretical studies, empirical experiments extensively explore the intrinsic link between the generalization performance of a model and loss landscapes. A consensus emerging from these studies, including \cite{chaudhari2019entropy, Jiang2020Fantastic, du2022sharpness, petzka2021relative}, is that flat minima usually yield better generalization compared to sharp minima.

\section{MIAdam}\label{sec:3}
In the parameter update formula of Adam, the first-order moment, as defined in Eq. (\ref{eq5a}), is reformulated as follows \cite{kingma2014adam}:
\begin{equation}
\begin{aligned}
\label{eq8}
m_{t,i}=&(1-\beta_{1})\sum^t_{j=0}\beta_{1}^{t-j}g_{j,i}.
\end{aligned}
\end{equation}
Consequently, the parameter update formula of Adam is rewritten as
\begin{equation}
\label{eq9}
\theta_{t+1,i}=\theta_{t,i}- \frac{\alpha (1-\beta_{1})\sum^t_{j=0}\beta_{1}^{t-j}g_{j,i}}{(1-\beta_1^t)\sqrt{\hat{v}_{t,i}+\epsilon}}.
\end{equation}
When the learning rate $\alpha$ is sufficiently small, Eq. (\ref{eq9}) is approximated in a continuous form as follows:
\begin{equation}
\begin{aligned}
\label{eq10}
\mathrm{d}\theta_{\tilde{t},i}&=\mu_{{\tilde{t}},i}\int_0^{\tilde{t}}\beta_{1}^{\tilde{t}-\tau} g_i (\tau) \mathrm{d}\tau,
\end{aligned}
\end{equation}
where ${\tilde{t}}$ is the continuous time, $\mathrm{d}\tau$ is equivalent to $\alpha$ and $\mu_{\tilde{t},i}=-(1-\beta_1)/((1-\beta_1^{\tilde{t}})\sqrt{\hat{v}_{{\tilde{t}},i}+\epsilon})$. It is noteworthy that the integral term appears in Eq. (\ref{eq10}).
In signal processing, a continuous input signal $x({\tilde{t}})$ that undergoes an integral operation is written as 
\begin{equation}
\begin{aligned}
\label{eq11}
y(\tilde{t})=\int_{{\tilde{t}}_0}^{{\tilde{t}}}x(\tau)\mathrm{d}\tau, \\
\end{aligned}
\end{equation}
where $y({\tilde{t}})$ is the integrated signal and the integration range is from ${\tilde{t}}_0$ to ${\tilde{t}}$. Ultimately, the resulting integral signal $y({\tilde{t}})$  contains the cumulative information of the original signal $x({\tilde{t}})$ at different points in time.  After the integral operation, the high-frequency components of the signal are filtered out. Inspired by this, the trajectory of the optimizer on the loss landscape can be viewed as the input signal when training a DNN, and the sharp minima are equivalent to the high-frequency components in the signal. Integrating this signal is equivalent to filtering out the sharp minima encountered by an optimizer in the loss landscape, thereby guiding the optimizer toward convergence in flat regions. Therefore, to further achieve the effect of filtering out the sharp minima encountered by the optimizer, multiple integrations, an enhanced version of the integral operation, are introduced into the parameter update formula of Adam.
Based on the process involving the integration as depicted in Eq. (\ref{eq11}), we obtain the following equation:
\begin{equation}
\begin{aligned}
\label{eq12}
&\mathrm{d}\theta_{{\tilde{t}},\mathcal{i}}=\\
&\mu_{{\tilde{t}},i}\!\!\overbrace{\int_0^{{\tilde{t}}}\! \!\!\kappa^{{\tilde{t}}\!-{\tilde{t}}_{1}}\!\!\!\int_0^{{\tilde{t}}_{1}}\!\!\!\!\!\!\cdots\!\!\int_0^{{\tilde{t}}_{n\!-\!2}}\!\!\!\kappa^{{\tilde{t}}_{n\!-\!2}\!-{\tilde{t}}_{n\!-\!1}}\!\!\!\int_0^{{\tilde{t}}_{n\!-\!1}}}^{\text{$n{\text{-th}}$-order multiple integration}}\!\!\!\beta_1^{{\tilde{t}}_{n\!-\!1}\!-\!\tilde{\tau}}\!\!\!\mathcal{g}_\mathcal{i}(\tilde{\tau})\mathrm{d}\tilde{\tau}\mathrm{d}{\tilde{t}}_{n\!-\!1}\!\cdots\!\mathrm{d}{\tilde{t}}_1,\\
\end{aligned}
\end{equation}
where $\kappa$ is multiple integration rate which adjusts the multiple integral term. Then, we perform cumulative operations on the first-order moments in the parameter update formula of Adam to transform the multiple integral term from a continuous form to its corresponding discrete form. Thus, the corresponding parameter update formula of Eq. (\ref{eq12}) is derived as follows:
\begin{equation}
\label{eq13}
\left\{
\begin{aligned}
&m_{t,i}=\beta_1 m_{t-1,i}+(1-\beta_1)g_{t,i},\\
&\overline{m}_{t,i}^{(n)}\!\!=\!(1\!-\!\beta_{1})\!\!\!\overbrace{\sum^t_{t_1=0}\!\!\kappa^{t-t_1}\!\!\!\!\sum^{t_1}_{t_2=0}\!\!\cdots\!\!\!\!\!\!\sum^{t_{n\!-\!2}}_{t_{n\!-\!1}=0}\!\!\!\!\kappa^{t_{n\!-\!2}-t_{n\!-\!1}}\!\!\!\!\sum^{t_{n\!-\!1}}_{t_{n}=0}\!\!\beta_1^{{t_{n\!-\!1}}\!-\!t_{n}}}^{\text{$n{\text{-th}}$-order multiple summation}}\!\!\!g_{t_{n},i}\\
 &\quad\ \!\ \ =\!\!\sum^t_{t_1=0}\!\kappa^{t-t_1}\!\!\sum^{t_1}_{t_2=0}\!\!\cdots\!\!\!\!\!\!\sum^{t_{n\!-\!2}}_{t_{n\!-\!1}=0}\!\!\!\!\kappa^{t_{n\!-\!2}-t_{n\!-\!1}}m_{t_{{n\!-\!1},i}},\\
&\theta_{t+1,i}=\theta_{t,i}- \frac{\alpha^n \overline{m}_{t,i}^{(n)} }{(1-\beta_1^t)\sqrt{\hat{v}_t+\epsilon}},
\end{aligned}
\right.
\end{equation}
where the superscript $^{(n)}$ means the $n$-th-order multiple summation. 
\begin{algorithm}[t]
\caption{MIAdam}
\label{alg: miadam}
\begin{minipage}{0.7\textwidth}
\textbf{Given:} Learning rate: $\alpha$;\\
\qquad exponential decay rates: $\beta_1$, $\beta_2$;\\
multiple integration rate: $\kappa$;\\
infinitesimal term: $\epsilon$; \\
the order of the multiple integral item: $n$; \\switching moment: $\zeta$.\\
\textbf{Initialize:} Step time $ t \gets 0$; \\first moment vector $m_{t=0,i}\gets0$; \\second moment vector $v_{t,i}\gets0$, $\overline{m}_{t=0,i}^{(0)}\gets m_{t=0,i}$.
\begin{algorithmic}[1]
\WHILE{stopping criterion is not met}
    \STATE $t \gets t+1$
    \STATE using Eq. (\ref{eq3}) to get the gradient $g_{t,i}$
    \STATE $m_{t,i} \gets \beta_1 m_{t-1,i} + (1-\beta_1)g_{t,i}$
    \STATE $v_{t,i} \gets \beta_2v_{t-1,i} + (1-\beta_2)g_{t,i}^2$
    \IF{$t<\zeta$}
    \STATE $m^{(0)}_{t,i} \gets m_{t,i}$
        \FOR{$j=1$ to $n$}
            \STATE $\overline{m}_{t,i}^{(j)}\gets\kappa\overline{m}_{t-1,i}^{(j)}+\overline{m}_{t,i}^{(j-1)}$
        \ENDFOR
        \STATE $\alpha_t \gets \alpha^n$
        \STATE $\hat{m}_{t,i} \gets \overline{m}_{t,i}^{(n)}/(1-\beta_1^t)$
    \ELSE  
        \STATE $\alpha_t \gets \alpha$
        \STATE $\hat{m}_{t,i} \gets m_{t,i}/(1-\beta_1^t)$
    \ENDIF
    \STATE $\hat{v}_{t,i} \gets v_{t,i}/(1-\beta_2^t)$
    \STATE $\theta_{t,i} \gets \theta_{t-1,i} - \alpha_t\hat{m}_{t,i}/(\sqrt{\hat{v}_{t,i}}+\epsilon)$
\ENDWHILE
\end{algorithmic}
\end{minipage}
\end{algorithm}
According to the theoretical analyses in Section \ref{sec:4} and the simulations in Fig. \ref{fig.2}, although the multiple integral term helps an optimizer to find flat minima,  the optimizer hovers around flat minima and does not converge. Thus, we only use the multiple integral term in the early stages of training, and after that, the optimizer switches to Adam to ensure that the training is convergent eventually. At this point, the multiple integral term is introduced into Adam, and this new optimizer is named MIAdam. The pseudo code for MIAdam is shown in Algorithm \ref{alg: miadam}.

Note that the multiple integration is approximated by the multiple summation in Algorithm \ref{alg: miadam}, which adds only $n$ additional summation operations at each iteration for each dimension of the parameter. Therefore, MIAdam adds very little additional computational overhead compared to Adam. In the following text, we refer to Adam with an additional first-order integration as MIAdam1, and the one with an additional second-order integration as MIAdam2, and so on.
\section{Generalization and Convergence Analyses}\label{sec:4}
In this section, we present the theoretical analyses of the generalization and convergence associated with the addition of the multiple integral term to Adam, which does not involve the switching of optimizers. These analyses provide a theoretical foundation for our proposed optimizer.
\subsection{Generalization Analyses}
In this subsection, the diffusion theory framework is utilized to rigorously demonstrate that the incorporation of the multiple integral term enhances the generalization capabilities of the model. Specifically, generalization is quantitatively assessed by comparing the mean escape time, represented as $\phi$, which indicates an optimizer's ability to escape from sharp minima. In the following analyses, we begin by delineating three fundamental assumptions that are crucial for the application of the diffusion theory framework \cite{xie2020diffusion}.
\begin{assumption}\label{asu:1}
  The loss function around the critical point $\bm{p}$ is approximately written as
\begin{equation}
 \label{eq:asu1}
  \begin{aligned}
L(\bm{\theta})=L(\bm{p})+\frac{1}{2}(\bm{\theta}-\bm{p})^\top H(\bm{p})(\bm{\theta}-\bm{p}),
\end{aligned}
\end{equation}
where the superscript $^{\top}$ means the transpose of a vector.
\end{assumption}
\begin{assumption}\label{asu:2}
(Quasi-equilibrium approximation). The system is in quasi-equilibrium near minima.
\end{assumption}
\begin{assumption}\label{asu:3}
(Low-temperature approximation). The system is under low temperature (small gradient noise).
\end{assumption}
 Consequently, following the theoretical analyses in  \cite{xie2020diffusion,xie2022adaptive}, we can further deduce Theorem \ref{thm:generalizaion}. The detailed proof is given in the Appendix.

\begin{theorem}\label{thm:generalizaion}
Suppose that Assumption \ref{asu:1}, Assumption \ref{asu:2}, and Assumption \ref{asu:3} hold while saddle point $\bm{u}$ is the exit from sharp minimum $\bm{a}$. Then the mean escape time of MIAdam1 from sharp minimum $\bm{a}$ to flat minimum $\bm{b}$ through saddle point $\bm{u}$ before the switch is
 \begin{equation}\label{eq1}
  \begin{aligned}
   &\phi_{\mathrm{MIAdam1}}=\\&\pi\left[\sqrt{1+\frac{4 \alpha \sqrt{\mathcal{b}\left|H_{{\bm{u}} {\bm{e}}}\right|}}{\tilde{t}(1-\beta_1)}}+1\right] \frac{\left|\operatorname{det}\left(H_{\bm{a}}^{-1} H_{\bm{u}}\right)\right|^{\frac{1}{4}}}{\left|H_{{\bm{u}} {\bm{e}}}\right|}\\
& \exp \left[\frac{2 \sqrt{\mathcal{b}} \Delta L}{\tilde{t}\alpha}\left(\frac{\varrho}{\sqrt{H_{{\bm{a}} {\bm{e}}}}}+\frac{(1-\varrho)}{\sqrt{\left|H_{{\bm{u}} {\bm{e}}}\right|}}\right)\right],
    \end{aligned}
\end{equation}
where subscript $_{\bm{e}}$ denotes the escape direction; $\varrho$ is the path-dependent parameter; $\mathcal{b} = |S|$ indicates the batch size; $\Delta L=L(\bm{u})-L(\bm{a})$; $H$ represents the Hessian matrix.
\end{theorem}
\begin{figure*}[t]\centering
    \subfigure[First Loss Landscape]
    {
        \includegraphics[width=0.235\textwidth]{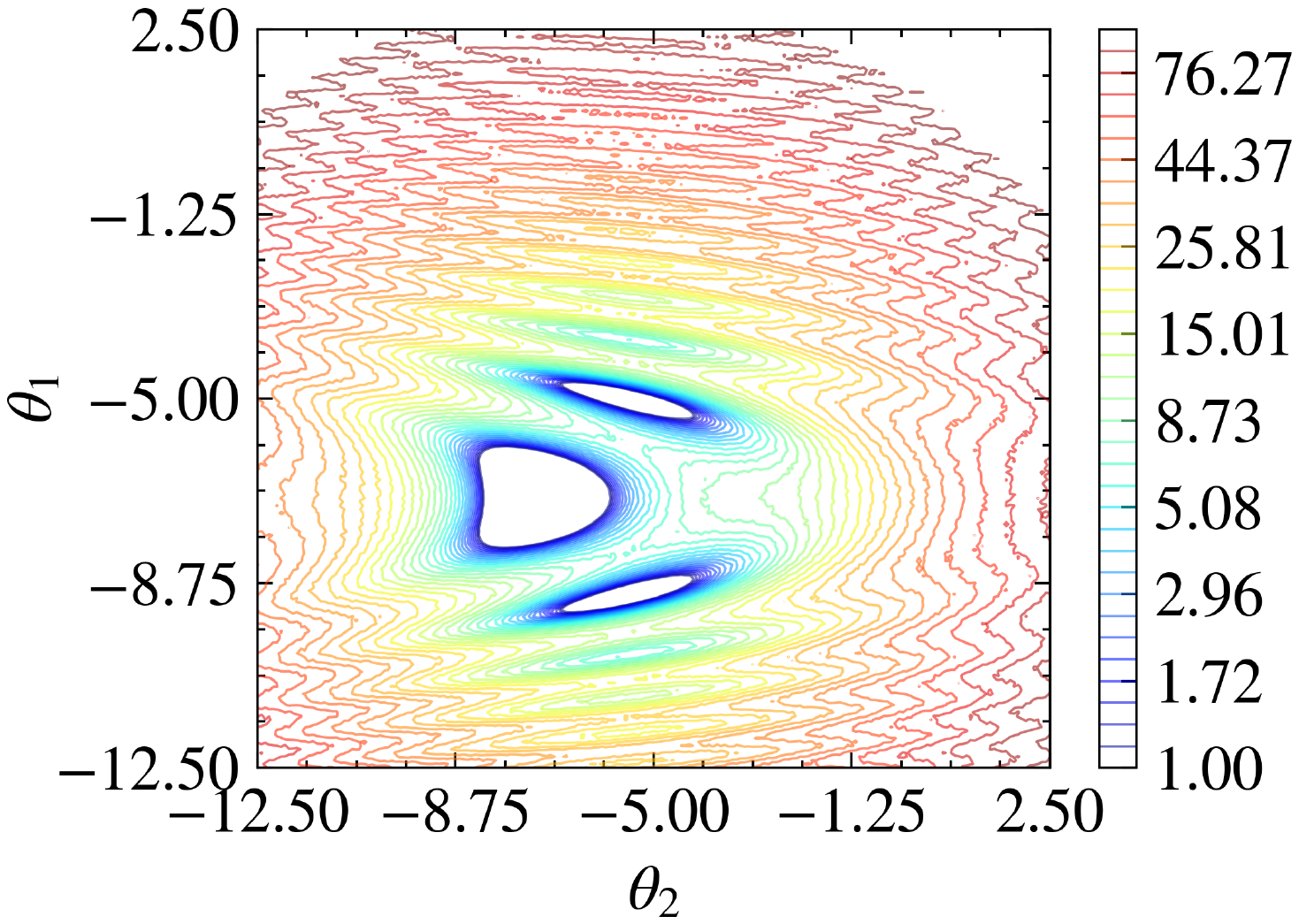}
    }\label{fig:2a}
    \subfigure[$\alpha$=0.05]
    {
        \includegraphics[width=0.22\textwidth]{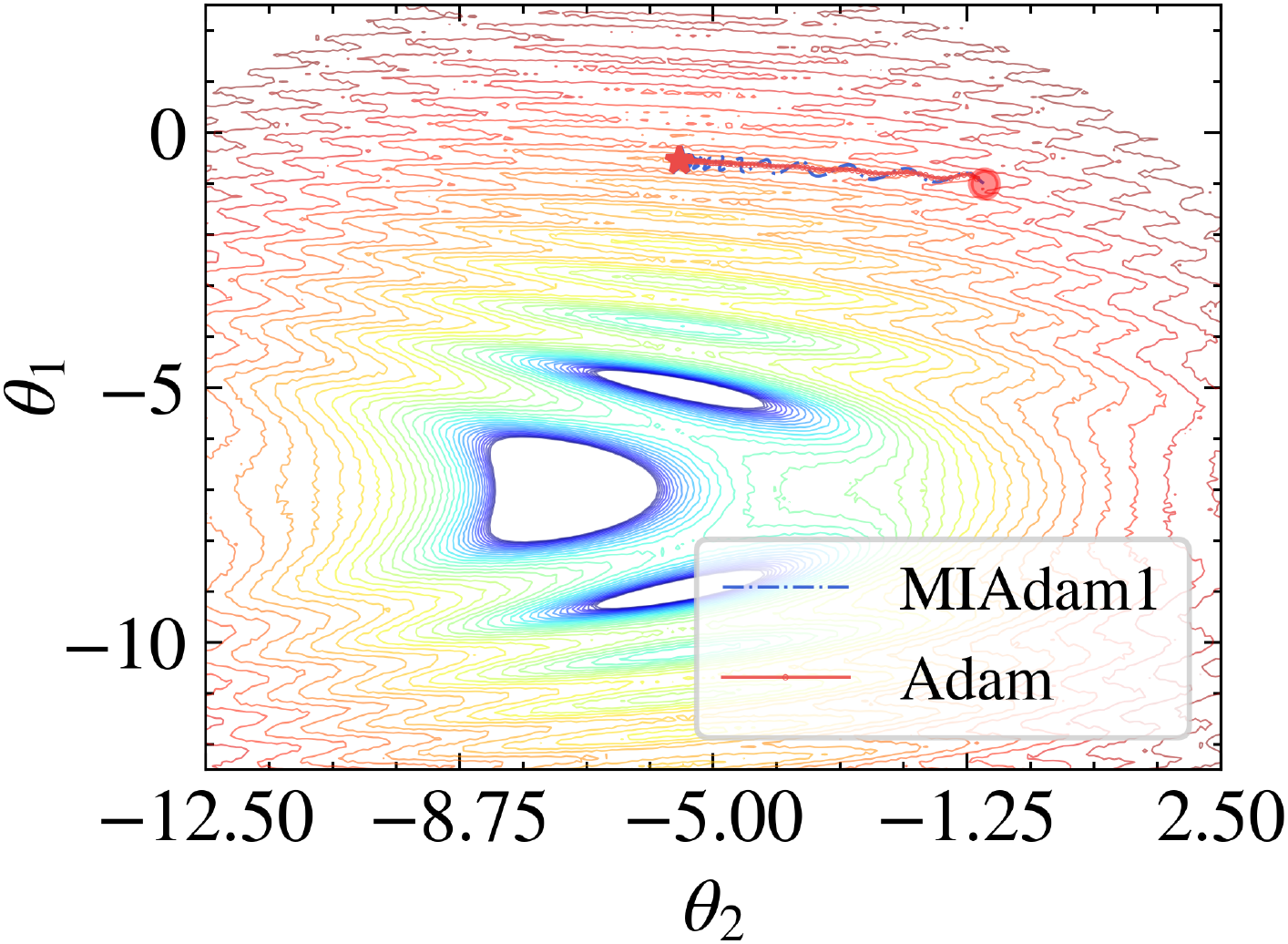}
    }\label{fig:2b}
    \subfigure[$\alpha$=0.1]
    {
        \includegraphics[width=0.22\textwidth]{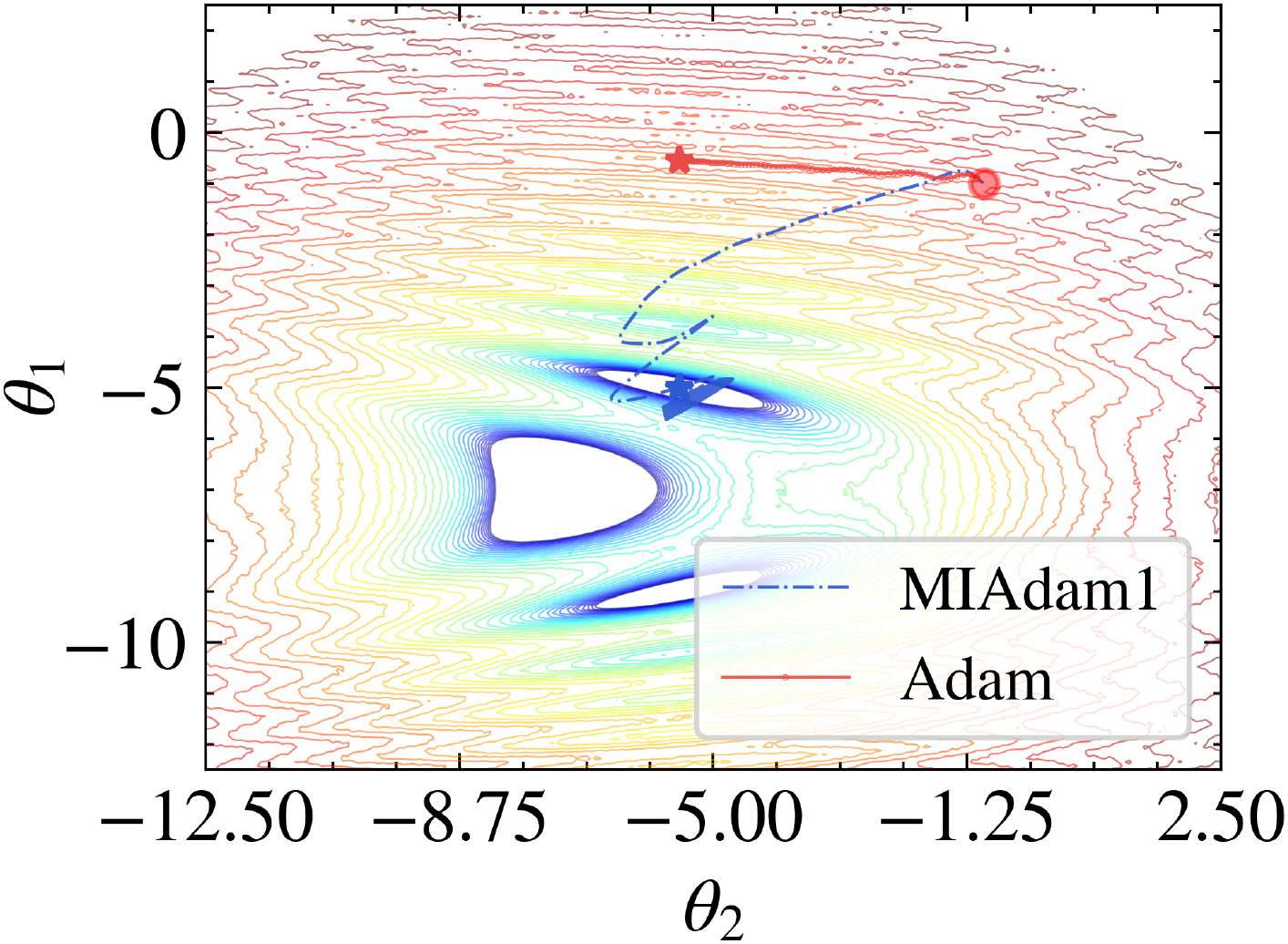}
    }\label{fig:2c}
    \subfigure[$\alpha$=0.15]
    {
        \includegraphics[width=0.22\textwidth]{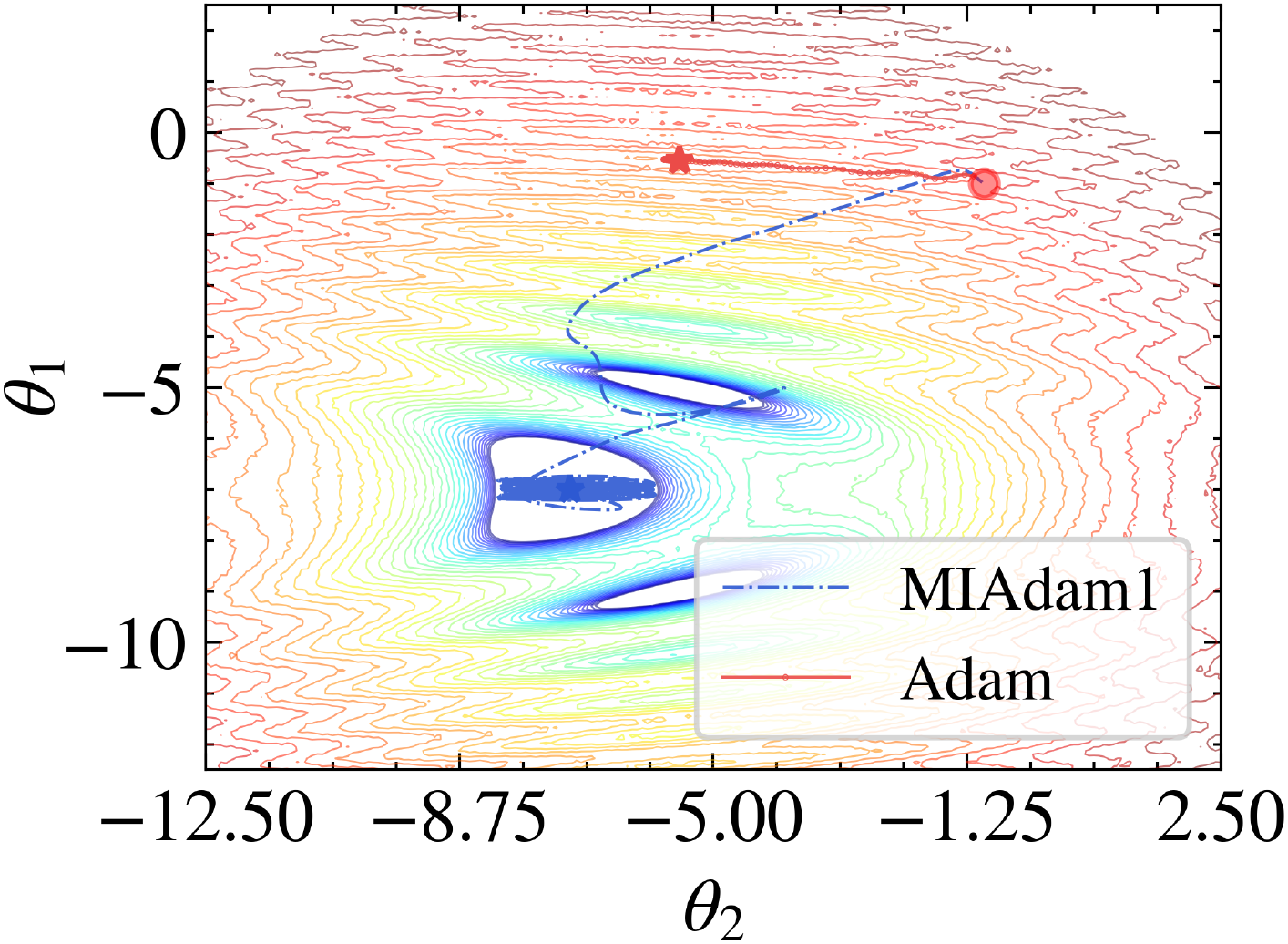}
    }\label{fig:2d}
 
   \hspace{0.25cm}\subfigure[Second Loss Landscape] 
    {
        \includegraphics[width=0.22\textwidth]{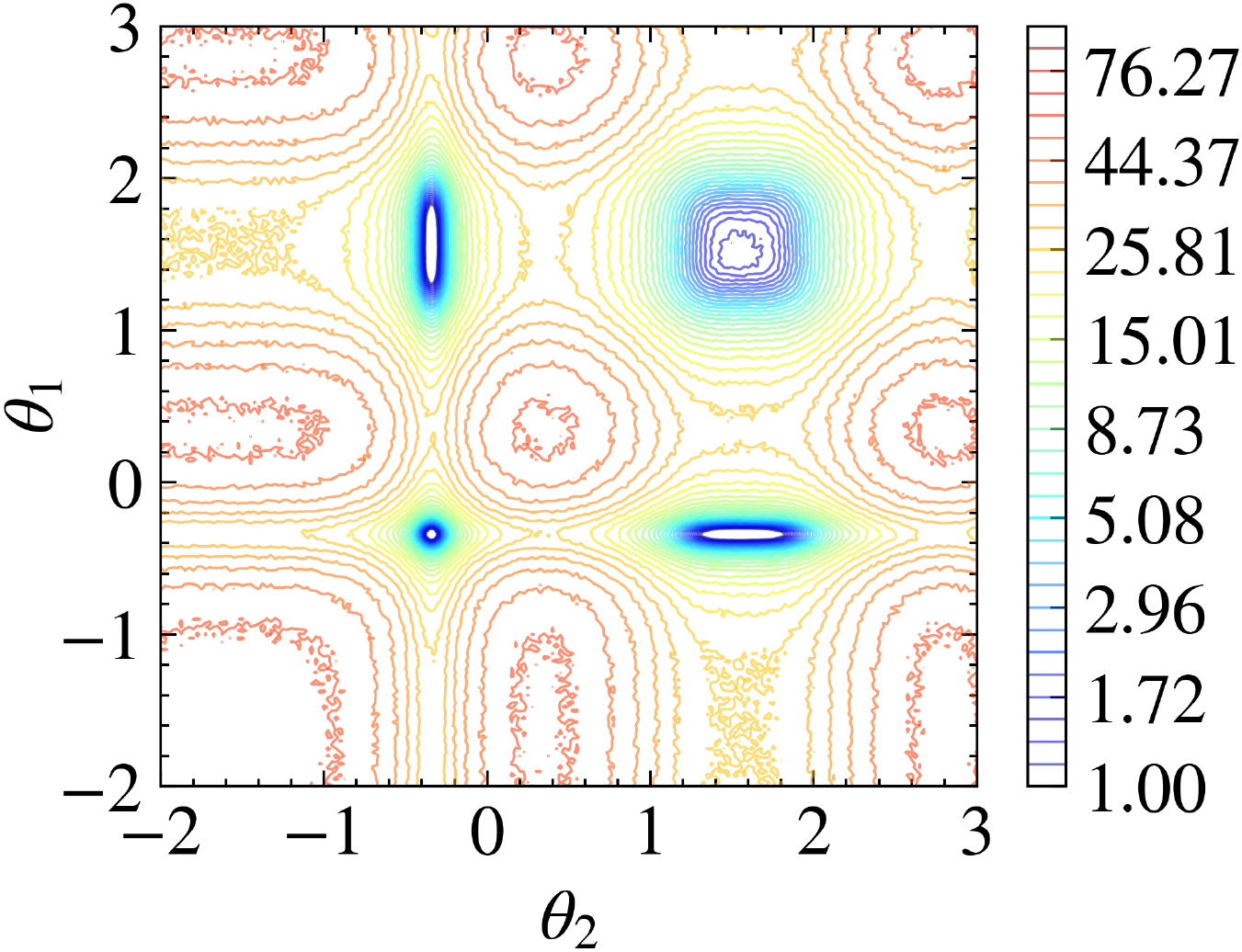}
    }\label{fig:2e}
    \hspace{0.08cm}\subfigure[MIAdam1 and Adam]
    {
        \includegraphics[width=0.205\textwidth]{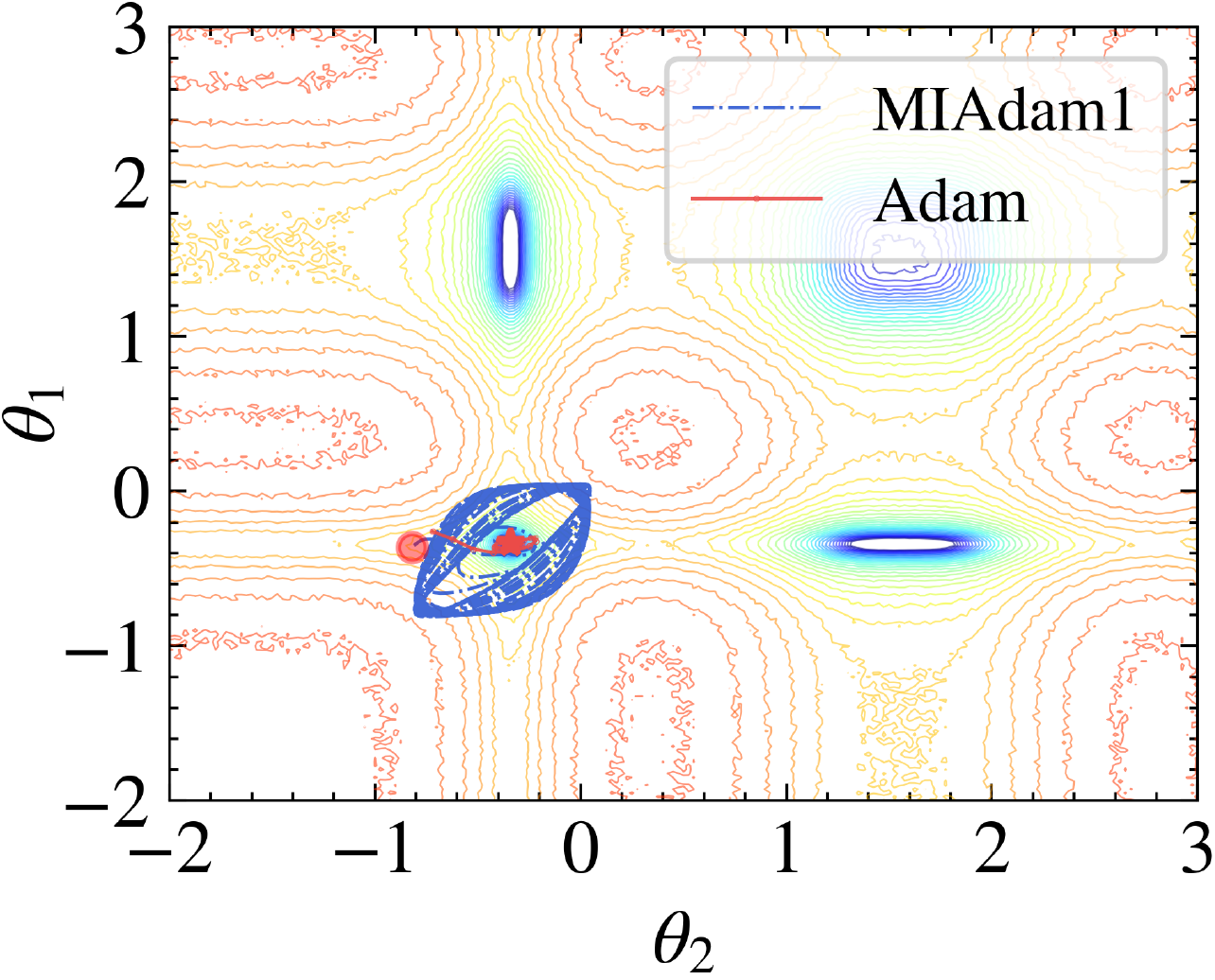}
    }\label{fig:2f}
    \hspace{0.25cm}\subfigure[MIAdam2 and Adam]
    {
        \includegraphics[width=0.205\textwidth]{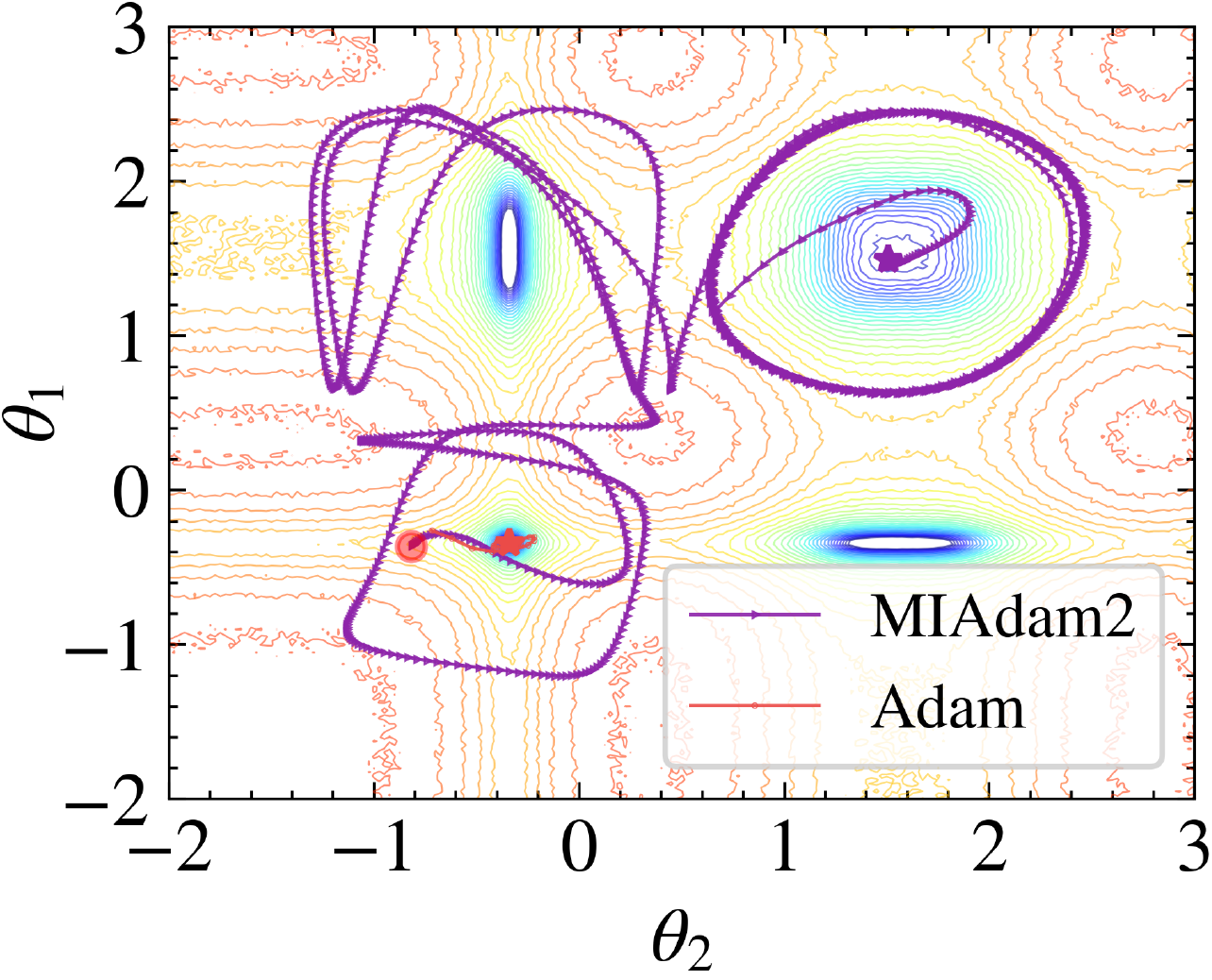}
    }\label{fig:2g}
     \hspace{0.255cm}\subfigure[MIAdam3 and Adam]
    {
        \includegraphics[width=0.205\textwidth]{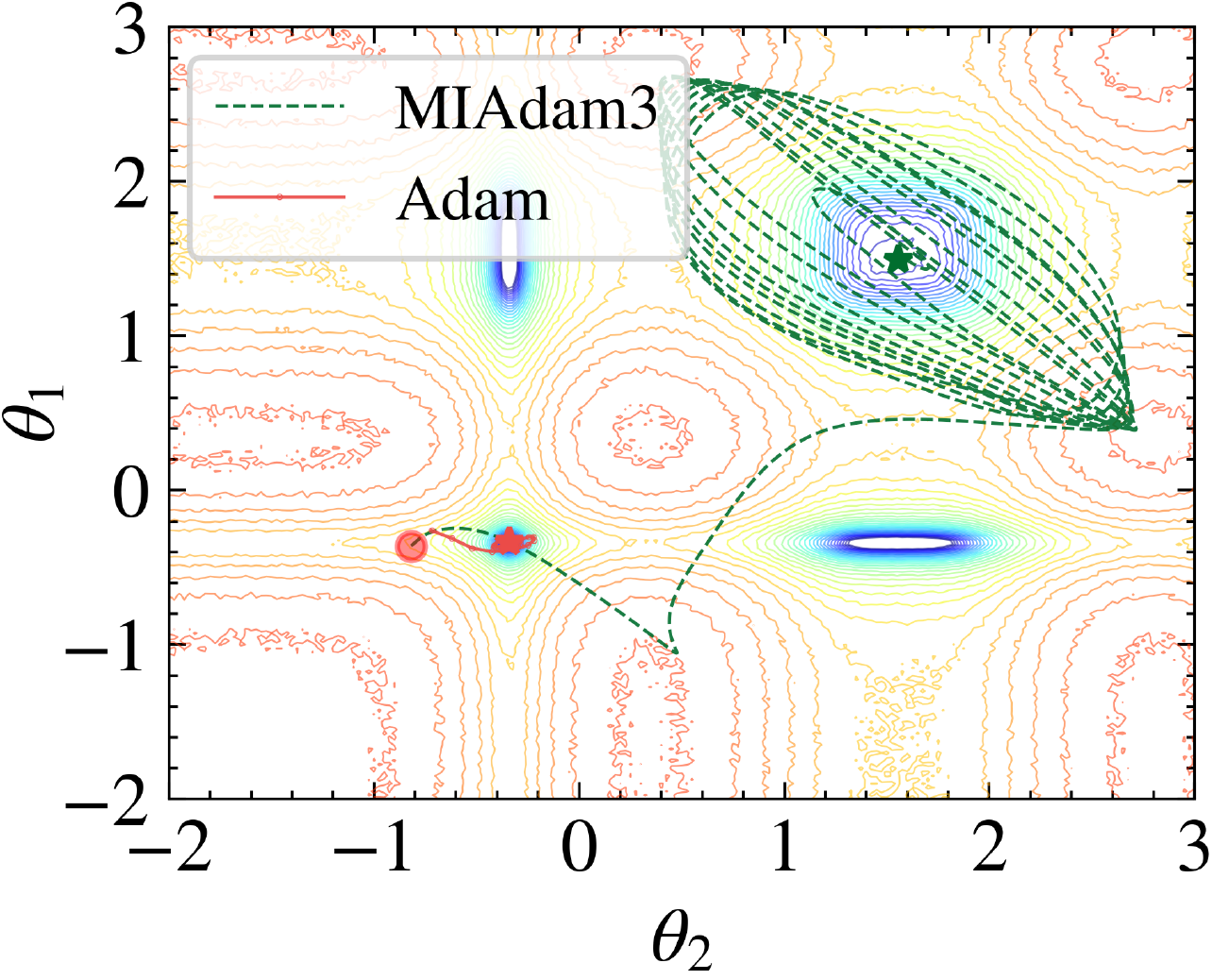}
    }\label{fig:2h}    
    \caption{Simulations of trajectory of Adam and MIAdam on 2-parameter loss landscapes. }
    \label{fig.2}
\end{figure*}
Comparing the mean escape time $\phi_{\mathrm{MIAdam1}}$ obtained from Theorem \ref{thm:generalizaion} with that of Adam's in  ~\cite{xie2022adaptive}, 
\begin{equation}
 \label{eq2}
\begin{aligned}
\phi_{\mathrm{Adam}}  
 =&\pi\left[\sqrt{1+\frac{4 \alpha \sqrt{\mathcal{b}\left|H_{\bm{u} \bm{e}}\right|}}{(1-\beta_1)}}+1\right] \frac{\left|\operatorname{det}\left(H_{\bm{a}}^{-1} H_{\bm{u}}\right)\right|^{\frac{1}{4}}}{\left|H_{{\bm{u}} {\bm{e}}}\right|}\\
 &\exp \left[\frac{2 \sqrt{\mathcal{b}} \Delta L}{\alpha}\left(\frac{\varrho}{\sqrt{H_{{\bm{a}} {\bm{e}}}}}+\frac{(1-\varrho)}{\sqrt{\left|H_{{\bm{u}} {\bm{e}}}\right|}}\right)\right],
\end{aligned}
\end{equation}
when $\tilde{t} > 1$, it is found that $\phi_{\mathrm{MIAdam1}}$ is smaller than $\phi_{\mathrm{Adam}}$, indicating that Adam introduces an additional first-order integration which is more likely to escape from sharp minima and consequently converge to flat minima, thereby improving the generalization. 
\subsection{Convergence Analyses}
In order to verify the effect of the multiple integral term on the convergence of the optimizer, we follow the analytical framework of Adam which is also used in this subsection. Concretely, the regret bound $R({\hat{t}})$ is utilized to evaluate the convergence of the algorithm and is defined as follows:
\begin{equation}\label{eq1}
  \begin{aligned}
        R({\hat{t}})=\sum_{t=1}^{\hat{t}} f_t(\bm{\theta}_{t})-\min _{\bm{\theta}} \sum_{t=1}^{\hat{t}} f_t(\bm{\theta}),
  \end{aligned}
\end{equation}
 where $f_t(\cdot)$ is a convex loss function. 
\begin{theorem} \label{thm:convergence}
 Assume that the convex function $f_{t}$ has bounded gradients, $\left\|\nabla f_{t}(\bm{\theta})\right\|_{2} \leq \mathsf{g},\left\|\nabla f_{t}(\bm{\theta})\right\|_{\infty} \leq$ $\mathsf{g}_{\infty}$ for all $\bm{\theta} \in \mathbb{R}^{d}$ and distance between any $\bm{\theta}_{t}$ is guaranteed to be bounded, $\left\|\bm{\theta}_{n}-\bm{\theta}_{m}\right\|_{2} \leq \mathsf{d}$, $\left\|\bm{\theta}_{m}-\bm{\theta}_{n}\right\|_{\infty} \leq \mathsf{d}_{\infty}$ for any $m, n \in\{1, \ldots, {\hat{t}}\}$, and $\beta_{1}, \beta_{2} \in[0,1)$ satisfy $\beta_{1}^{2}/\sqrt{\beta_{2}}<1$. Let $\alpha_{t}=\alpha/t^h$ and $\beta_{1, t}=\beta_{1} \lambda^{t-1},\kappa \in(0,1], \lambda \in(0,1)$. For the convex problem, the  $R({\hat{t}})$ of MIAdam1 before the switch satisfies
 \begin{equation}\label{eq1}
  \begin{aligned}
\lim_{{\hat{t}}\rightarrow \infty}\frac{R({\hat{t}})}{{\hat{t}}}\neq 0.
  \end{aligned}
\end{equation}
\end{theorem}
The detailed proof of Theorem \ref{thm:convergence} is thoroughly presented in the Appendix. From Theorem \ref{thm:convergence}, it is evident that merely adding an extra first-order integration to the parameter updating formula of Adam leads to the non-convergence of the optimizer. Although it is non-convergent, it effectively escapes sharp minima and hovers around flat minima in the loss landscape. This observation is corroborated by the simulation results shown in Figs. \ref{fig.2}(f)-(h). As a result, the MIAdam's algorithm is structured to switch to Adam after a certain number of epochs to guarantee convergence.

\section{Simulations and Experiments}\label{sec:5}
In this section, we conduct the simulations on 2-parameter loss landscapes to illustrate the efficiency of MIAdam to escape from sharp minima. Furthermore, extensive empirical experiments are conducted to demonstrate that MIAdam outperforms Adam in terms of generalization and robustness against label noises. 

  \begin{figure*}[t]\centering
    \subfigure[Adam]
    {
        \includegraphics[width=0.21\textwidth]{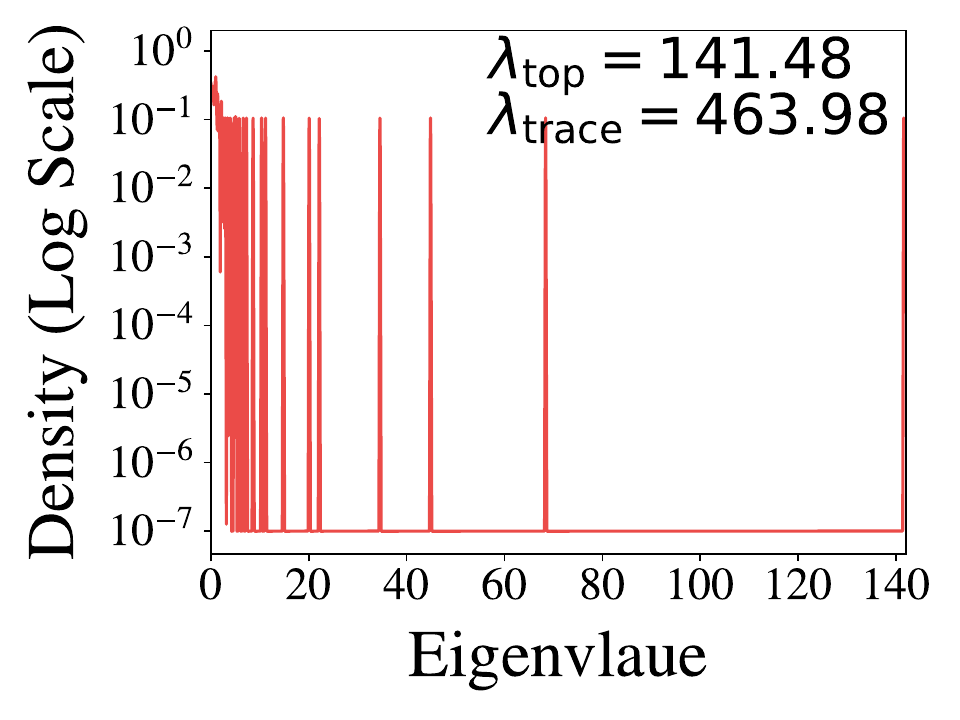}
    }
    \subfigure[MIAdam1]
    {
        \includegraphics[width=0.21\textwidth]{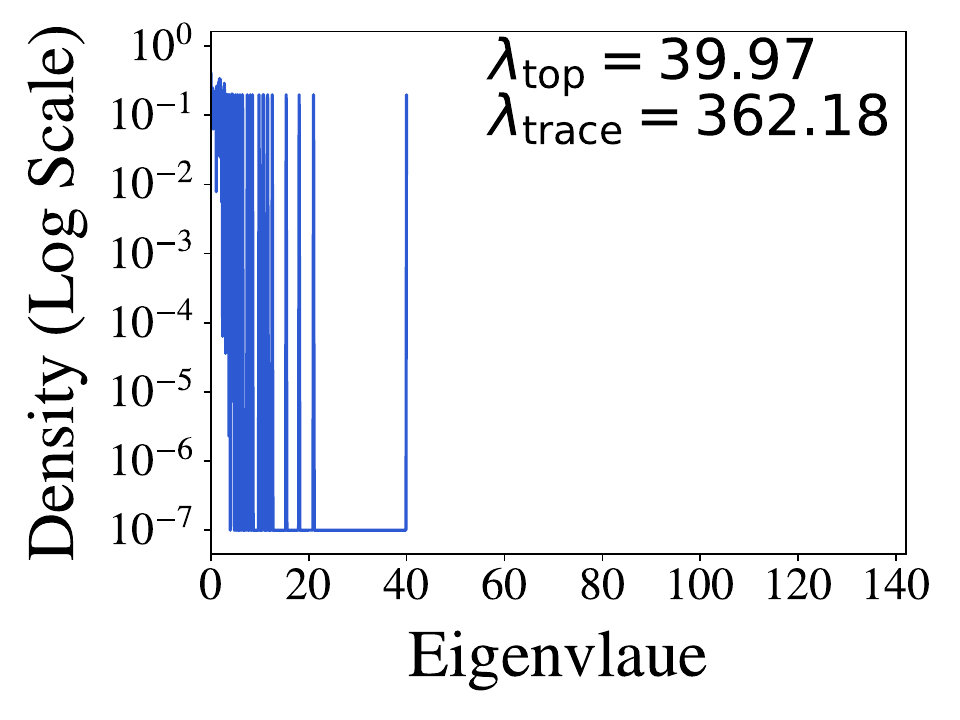}
    }
    \subfigure[MIAdam2]
    {
        \includegraphics[width=0.21\textwidth]{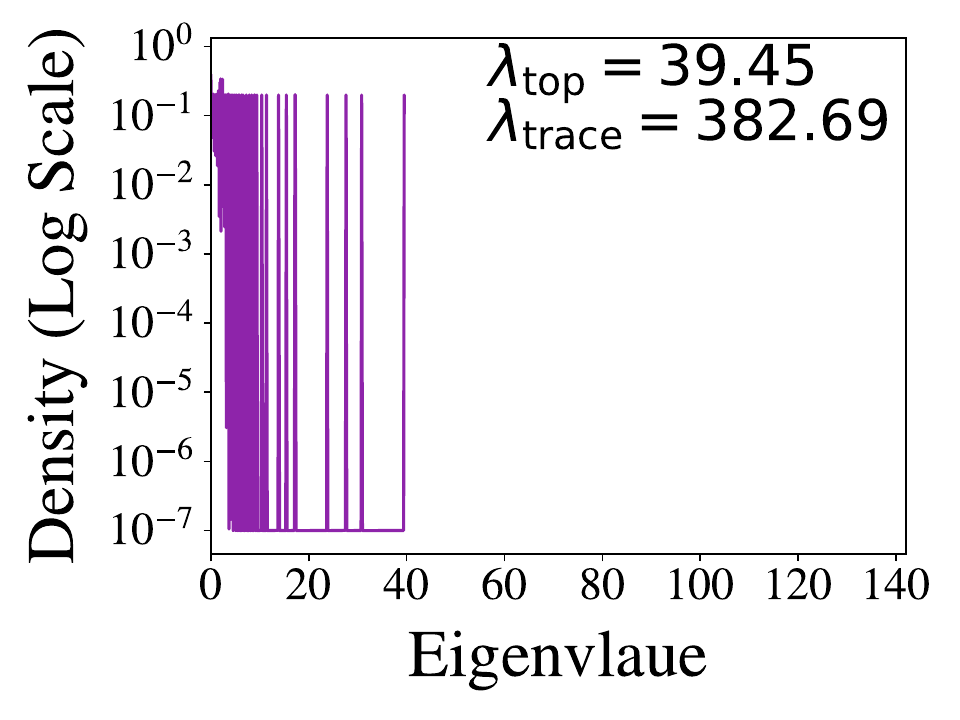}
    }
    \subfigure[MIAdam3]
    {
        \includegraphics[width=0.21\textwidth]{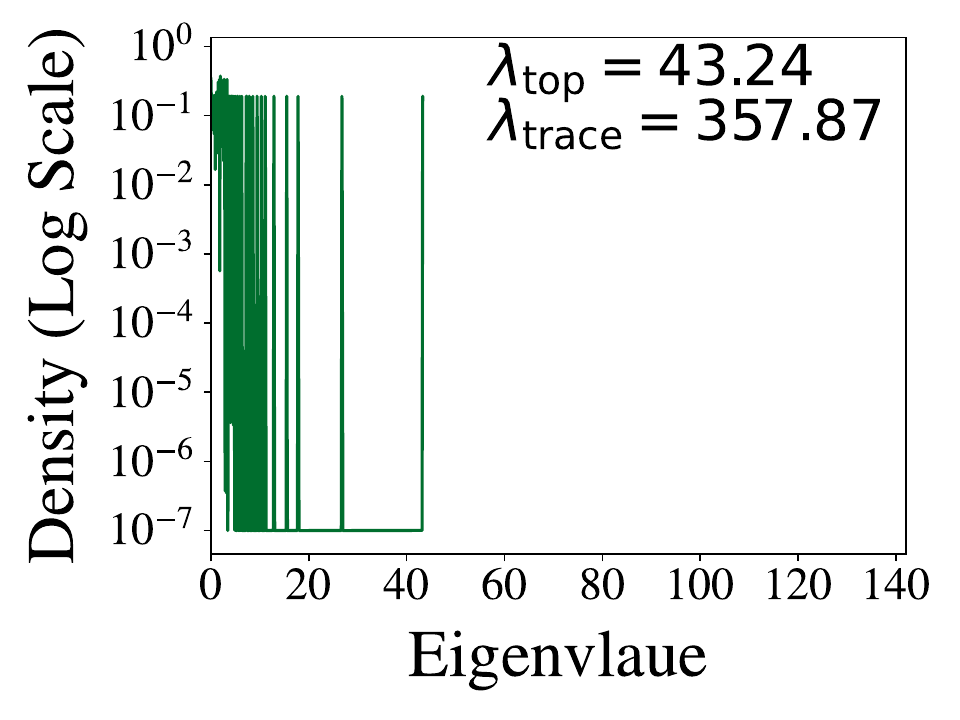}
    }
 
    \caption{Comparisons of top Hessian eigenvalues $\lambda_{\text{top}}$,
Hessian traces $\lambda_{\text{trace}}$, and full Hessian eigenvalue densities for loss landscapes on the CIFAR-10 dataset using ResNet18.}
    \label{fig.3}
\end{figure*}
\begin{table*}[!h]
    \renewcommand\arraystretch{1.1}
    \centering
    \setlength{\tabcolsep}{1mm}{
    \begin{tabular}{l|l|l|l|l|l|l|l|l}
        \toprule
        \textbf{Optimizer} & \multicolumn{4}{c|}{\textbf{ResNet18}} & \multicolumn{4}{c}{\textbf{ResNet50}} \\
        & \textbf{CIFAR-10(\%)}& \textbf{Time}& \textbf{CIFAR-100(\%)} & \textbf{Time} & \textbf{CIFAR-10(\%)} & \textbf{Time}& \textbf{CIFAR-100(\%)} & \textbf{Time} \\
        \hline
        Adam & $93.89_{\pm 0.18}$  & 47m & $73.17_{\pm0.26}$ & 47m & $93.69_{\pm 0.28}$ & 2h 45m& $75.24_{\pm0.62}$ & 2h 47m \\
        NAdam & $93.92_{\pm 0.11}$  &46m & $73.08_{\pm0.31}$ & 46m & $94.03_{\pm 0.07}$ & 2h 57m& $74.95_{\pm 0.08}$ & 2h 58m \\
        AdamW & $93.74_{\pm 0.07}$  & 48m & $72.11_{\pm0.23}$ & 47m & $93.80_{\pm 0.11}$ & 2h 45m & $74.40_{\pm0.61}$ & 2h 48m \\
        ND-Adam & $93.69_{\pm 0.08}$ & 44m& $72.26_{\pm0.34}$ & 48m & $93.81_{\pm 0.18}$ & 2h 39m& $74.18_{\pm0.35}$ & 2h 57m \\
        Adamax & $93.68_{\pm 0.25}$ & 1h 28m& $74.27_{\pm0.19}$ & 1h 27m & $94.10_{\pm 0.19}$ & 2h 48m&$76.30_{\pm0.10}$ & 2h 52m \\
        AdaBound & $92.93_{\pm 0.11}$ & 48m& $72.51_{\pm0.23}$ & 49m & $93.14_{\pm 0.29}$ & 3h 1m& $73.87_{\pm0.27}$ & 3h 2m \\
        SWATS & $92.73_{\pm 0.10}$ & 1h 6m& $72.76_{\pm0.27}$ & 54m & $92.38_{\pm 1.38}$ & 2h 51m& $72.83_{\pm2.10}$ & 2h 50m \\
        Adai & $93.86_{\pm 0.15}$ & 53m& $74.82_{\pm 0.09}$ & 53m & $93.95_{\pm 0.06}$ & 3h 32m& $76.07_{\pm0.36}$ & 3h 18m \\
        MIAdam1 & \bm{$94.20_{\pm 0.12}$}* & 47m& \bm{$75.03_{\pm0.05}$}* & 47m & \bm{$94.17_{\pm 0.10}$} & 2h 49m& \bm{$76.96_{\pm0.31}$}* & 2h 47m \\
        MIAdam2 & \bm{$94.03_{\pm 0.12}$} & 47m& \bm{$74.41_{\pm0.38}$} & 46m & $\bm{94.28_{\pm 0.15}}$* & 2h 47m& \bm{$76.63_{\pm0.15}$} & 2h 46m \\
        MIAdam3 & \bm{$93.96_{\pm 0.17}$} & 47m& \bm{$74.53_{\pm0.22}$} & 47m & $\bm{94.10_{\pm 0.10}}$ & 2h 48m& \bm{$76.51_{\pm0.02}$} & 2h 46m \\
        \hline

        \textbf{} & \multicolumn{4}{c|}{\textbf{DenseNet121}} & \multicolumn{4}{c}{\textbf{PyramidNet110}} \\
        & \textbf{CIFAR-10(\%)} & \textbf{Time }& \textbf{CIFAR-100(\%)} & \textbf{Time } & \textbf{CIFAR-10(\%)} & \textbf{Time }& \textbf{CIFAR-100(\%)} & \textbf{Time } \\
        \hline
        Adam & $94.11_{\pm 0.01}$ & 3h 30m& $73.88_{\pm0.34}$ & 3h 32m & $93.45_{\pm 0.10}$ & 2h 46m& $72.20_{\pm0.09}$ & 2h 50m\\
        NAdam & $94.39_{\pm 0.30}$ & 4h 23m& $65.71_{\pm0.34}$ & 4h 24m & $93.33_{\pm 0.20}$ & 3h 34m& $71.26_{\pm0.44}$ & 3h 32m \\
        AdamW & $94.16_{\pm 0.11}$ & 3h 25m& $75.07_{\pm0.15}$ & 3h 30m & $93.25_{\pm 0.08}$ & 2h 48m& $70.41_{\pm0.69}$ & 2h 43m \\
        ND-Adam & $94.11_{\pm 0.01}$ &3h 23m & $74.59_{\pm0.20}$ & 3h 34m & $93.00_{\pm 0.15}$ &2h 58m& $70.72_{\pm0.30}$ & 3h 11m \\
        Adamax & $90.97_{\pm 0.15}$ & 3h 40m& $63.48_{\pm0.08}$ & 3h 47m & $92.46_{\pm 0.24}$ & 4h 51m& $69.43_{\pm0.28}$ & 5h 16m\\
        AdaBound & $93.14_{\pm 0.10}$ & 4h 0m& $73.88_{\pm0.35}$ & 4h 2m & $92.14_{\pm 0.22}$ & 3h 32m& $68.97_{\pm0.39}$ & 3h 27m \\
        SWATS & $93.64_{\pm 0.94}$ & 3h 35m& $75.62_{\pm3.10}$ & 3h 45m & $89.42_{\pm 4.13}$ & 3h 35m& $49.82_{\pm25.13}$ & 2h 29m \\
        Adai & $94.45_{\pm 0.21}$ & 4h 28m& $76.77_{\pm0.31}$ & 4h 42m & $93.50_{\pm 0.10}$ & 4h 18m& $71.94_{\pm0.31}$ & 4h 5m \\
        MIAdam1 & \bm{$94.75_{\pm 0.10}$}* & 3h 29m& \bm{$77.02_{\pm0.10}$}* & 3h 26m & \bm{$93.65_{\pm 0.08}$}* & 2h 59m& \bm{$72.51_{\pm0.24}$}* & 2h 56m \\
        MIAdam2 & \bm{$94.43_{\pm 0.12}$} & 3h 29m& \bm{$76.21_{\pm0.33}$} & 3h 32m &$93.02_{\pm0.10}$ & 2h 52m& $71.96_{\pm0.59}$ & 2h 52m \\
        MIAdam3 & \bm{$94.35_{\pm0.03}$} & 3h 30m& \bm{$76.54_{\pm0.28}$} & 3h 30m & $93.07_{\pm 0.25}$ & 2h 54m& $71.34_{\pm0.17}$ & 2h 53m \\
        \bottomrule
    \end{tabular}
}    \caption{Top-1 test accuracy (mean$\pm$std) on CIFAR-10 and CIFAR-100.}
    \label{tab: cifar}
\end{table*}
\subsection{Simulations}
This subsection mainly includes simulations demonstrating that MIAdam is easier to escape from a sharp minima compared to Adam and exploring the impact of learning rate on the optimization process.
The first simulation is conducted on an elaborate 2-parameter loss landscape \cite{yang2020stochastic} with one flat minima surrounded by two sharp minima, whose contour map is displayed in Fig. \ref{fig.2}(a). On this loss landscape, the learning rates of Adam and MIAdam1 are respectively set to 0.05, 0.1, and 0.15, and the simulation results for their corresponding optimization trajectories are shown in Figs. \ref{fig.2}(b)-(d).  It is clear that MIAdam1 tends to escape from sharp minima and converge to flat minima compared to Adam on the 2-parameter loss landscape. Moreover, as the learning rate increases, MIAdam1 is able to converge to the flat minima. In contrast, Adam always shows poor convergence on this loss landscape and can not converge well to the flat minima or sharp minima. Therefore, our proposed method is effective in finding flat minima and can not be simply replaced by increasing the learning rate of Adam. 

The second 2-parameter loss landscape used for simulations is depicted in Fig. \ref{fig.2}(e), which contains a large number of sharp minima and flat minima. On this loss landscape, the optimization trajectories of MIAdam1, MIAdam2, MIAdam3, and Adam are compared in Fig. \ref{fig.2}(f)-(h), respectively. Simulation results indicate that MIAdam1, MIAdam2, and MIAdam3 tend to converge toward flat minima, while the Adam optimizer tends to converge to the nearest sharp minima. It's worth noting that MIAdam3 exhibits more intense oscillations near the flat region compared to the MIAdam2. This suggests that increasing the order of multiple integration does not always lead to improved outcomes. Different starting points influence the trajectory of the optimizer. Therefore, to make the simulation results more convincing, we conduct additional simulations using 2,500 different starting coordinate points and calculate the sum of the absolute values of the eigenvalues of the Hessian matrix of the different optimizers for the final convergence points at different starting coordinate points and compare them. Due to space constraints, the simulation results are presented in the Appendix.

\subsection{Experiments}
The effectiveness of MIAdam is evaluated in this subsection through extensive empirical experiments. Initially, we conduct image classification experiments with various neural network architectures on CIFAR\footnote{http://www.cs.toronto.edu/~kriz/cifar.html} and ImageNet-1k\footnote{https://www.image-net.org/}, compared against widely-used adaptive learning rate optimizers, including Adam and its SOTA variants. Additionally, we utilize the fast computation method of Hessian information of loss landscapes provided in \cite{yao2020pyhessian} for further comparative analyses. Subsequently, the effectiveness of the proposed MIAdam optimizer for text classification tasks is tested using the BERT and RoBERTa models across four distinct datasets \cite{lin2021bertgcn}. Finally, to validate the robustness against label noises of MIAdam, we perform image classification experiments on datasets injected with label noises. 
The results of MIAdam exceeding Adam are all bold, and the optimal experimental results are all marked by asterisks.
Because of space constraints, the detailed experimental settings for all experiments are included in the Appendix. 

\subsubsection{Image Classification Experiments}
To enhance the conviction of our experimental results, we employ four different neural network architectures for image classification tasks on the CIFAR-10 and CIFAR-100 datasets:  ResNet18 \cite{he2016deep}, ResNet50 \cite{he2016deep}, DenseNet121 \cite{Huang_2017_CVPR}, and PyramidNet110 \cite{han2017deep}. For experiments on large-scale image datasets, we utilize the AlexNet \cite{krizhevsky2012imagenet}, ResNet18, and DenseNet121 architectures for both training and testing on the ImageNet-1k. The classification performance of MIAdam is compared with optimizers such as Adam, NAdam \cite{dozat2016incorporating}, AdamW \cite{loshchilov2017decoupled}, ND-Adam, Adamax \cite{kingma2014adam}, AdaBound, SWATS, and Adai \cite{xie2022adaptive}. Detailed hyperparameters and experimental settings are presented in the Appendix. As observed from Table \ref{tab: cifar} and Table \ref{tab:image}, MIAdam maintains a training time comparable to Adam while obtaining much better performance than Adam. To provide a comparison of the flatness in the final convergence regions, we compute top Hessian eigenvalues,
Hessian traces, and full Hessian eigenvalue densities for loss landscapes of Adam, MIAdam1, MIAdam2, and MIAdam3 using DenseNet121 on the CIFAR-100 dataset in Fig. \ref{fig.3}. Fig. \ref{fig.3} suggests that the multiple integral term is helpful in finding flatter minima in a specific neural network training task. 

\begin{table}[t]
   \renewcommand\arraystretch{1.1}
   \centering
   \setlength{\tabcolsep}{1mm}{
       \begin{tabular}{llll}
           \toprule 
           \textbf{Optimizer}  & \textbf{AlexNet}(\%) & \textbf{ResNet18}(\%) & \textbf{DenseNet121}(\%) \\
           \midrule
           Adam    & $46.48$ & $67.19$ & $71.48$ \\
           MIAdam1 & \bm{$52.34$}* & \bm{$72.27$}* & \bm{$75.39$}*  \\
           MIAdam2 & \bm{$49.61$} & $70.70$ & \bm{$74.22$}  \\
           MIAdam3 & $43.36$ & \bm{$70.31$} & $\bm{74.60}$  \\
           \bottomrule
       \end{tabular}
   }
   
    \caption{Top-1 test accuracies (mean$\pm$std) on ImageNet-1k}\label{tab:image}
\end{table}

\begin{table}[h]
   \renewcommand\arraystretch{1.1}
   \centering
      \setlength{\tabcolsep}{1mm}{
   \begin{tabular}{llll}
    \multicolumn{2}{l}{}  \\
    \toprule 
    \textbf{Dataset} & \textbf{Optimizer}  & \textbf{BERT}(\%) &  \textbf{RoBERTa}(\%)
    \\
    \hline
       \multirow{4}*{R8}
       & Adam    & $98.15_{ \pm 0.02}$ & $98.36_{ \pm 0.05}$\\
       & MIAdam1 & $\bm{98.18_{ \pm 0.03}}$* & $\bm{98.45_{ \pm 0.05}}$* \\
       & MIAdam2 & $98.11_{ \pm 0.10}$ & $98.29_{\pm0.10}$ \\
       & MIAdam3 & $98.04_{\pm0.09}$ & $98.29_{\pm0.06}$ \\
    \hline
       \multirow{4}*{R52}
       & Adam    & $96.36_{ \pm 0.21}$ & $96.21_{ \pm 0.13}$ \\
       & MIAdam1 & \bm{$96.42_{ \pm 0.23}$} & \bm{$96.48_{ \pm 0.01}$} \\
       & MIAdam2 & $\bm{96.57_{ \pm 0.07}}$ & $\bm{96.46_{\pm0.23}}$ \\
       & MIAdam3 & $\bm{96.65_{\pm0.27}}$* & $\bm{96.65_{\pm0.07}}$* \\
     \hline
       \multirow{4}*{MR}
       & Adam    & $86.03_{ \pm 0.30}$ & $87.72_{ \pm 0.09}$ \\
       & MIAdam1 & \bm{$86.51_{ \pm 0.09}$}* & \bm{$89.73_{ \pm 0.17}$}* \\
       & MIAdam2 & \bm{$86.45_{ \pm 0.19}$} & \bm{$89.54_{\pm0.19}$} \\
       & MIAdam3 & $\bm{86.34_{\pm0.09}}$ & \bm{$89.54_{\pm0.18}$} \\ 
   \bottomrule
   \end{tabular}
   }     \caption{Test accuracies (mean$\pm$std) on text classification experiments}\label{tab:text}
\end{table}
\subsubsection{Text Classification Experiments}
We conduct text classification experiments by fine-tuning the pre-trained models, BERT and RoBERTa models, on three widely-used text datasets R8, R52, and Movie Review (MR). Each optimizer is run three times on each dataset using different network structures, with the mean and standard deviation of the test accuracy reported in Table \ref{tab:text}. The experimental results indicate that MIAdam significantly outperforms Adam in text classification tasks.
\subsubsection{Robustness Against Label Noises}
In this subsection, we investigate the capacity of MIAdam to withstand label noises in the training dataset, thereby validating its robustness against label noises.
The ResNet18 network is trained by using Adam and MIAdam on the corrupted version of the CIAFR10 dataset, where some of its training labels are randomly flipped while the inputs are kept clean. The noise levels are 20\%, 40\%, 60\%, and 80\%. On each noise level, each optimizer is run only once.
The remaining experimental settings are consistent with those used in the previous image classification experiments. As indicated in Table \ref{tab:noise}, MIAdam consistently achieves the highest test accuracy across all noise levels, underscoring MIAdam's superior robustness against label noises.
\begin{table}[h]
   \renewcommand\arraystretch{1.1}
   \centering
   \setlength{\tabcolsep}{1mm}{
   \begin{tabular}{llllll}
    \multicolumn{2}{l}{}  \\
    \toprule 
     \multirow{2}*{\textbf{Optimizer}}  & \multicolumn{4}{c}{\textbf{Noise rate}(\%)}   \\
 & 20& 40& 60 & 80     \\
    \hline     
        Adam    & $88.24$ & $84.90$ &$79.61$ &$66.39$\\
        ND-Adam    & $87.52$ & $84.10$ &$78.34$ &$63.51$\\        
        AdaBound    & $86.51$ & $82.46$ &$76.58$ &$57.86$\\
        SWATS    & $89.43$ & $85.47$ &$80.30$ &$53.50$\\ 
        Adai    & $86.09$ & $81.92$ &$75.72$ &$58.60$\\      
        MIAdam1 & \bm{$90.32$}* &\bm{$87.67$}* &\bm{$82.02$}*  &\bm{$67.68$}*      \\
        MIAdam2 & \bm{$89.13$} &\bm{$85.03$} &$\bm{79.84}$  &$64.96$    \\
        MIAdam3 & \bm{$88.71$} & \bm{$85.87$} &$79.40$ &$64.27$     \\
   \bottomrule
   \end{tabular}
   }      
   \caption{Top-1 test accuracy on CIFAR-10 under label noises.}
   \label{tab:noise}
\end{table}
\section{Conclusion}
In this paper, we have proposed MIAdam, a new adaptive learning rate optimizer algorithm with a multiple integral term added to Adam. MIAdam smoothes the optimization trajectory through the filtering effect of the multiple integral term, enabling it to escape sharp local minima during training and converge towards flat minima, thereby alleviating the problem of poor generalization of Adam and improving the robustness against label noises while retaining the fast convergence of Adam. Utilizing the diffusion theory framework, we have provided the proof that incorporating the multiple integral term enhances the capability of the optimizer to escape sharp minima and converge to flatter minima, thus improving the generalization of the models. We have analyzed the convergence of MIAdam and provided a guarantee of convergence.  The simulations have demonstrated that MIAdam is capable of finding flatter minima compared to Adam. For empirical analyses, We have conducted image classification experiments, text classification experiments, and experiments that inject label noises into datasets. The experimental results show that MIAdam has much better generalization and robustness against label noises than Adam. Future work will focus on introducing multiple integral terms into other optimizers.

\section{Acknowledgments}
This work was supported in part by the National Natural Science Foundation of China under Grant 62476115 and Grant 62176109, in part by the Fundamental Research Funds for the Central Universities under Grant lzujbky2023-ct05 and Grant Izuibky-2023-ey07, in part by the China Computer Federation (CCF)-Baidu Open Fund under Grant 202306, and in part by the Supercomputing Center of Lanzhou University.

\bibliography{aaai25}

\newpage
\appendix
\onecolumn
\section{Appendix}
\section{Simulation and Experiment Settings}\label{appendix-experiment}
All experimental procedures are implemented using Python 3.6 and PyTorch 1.8.3. Computational tasks are executed on a system running Ubuntu 16.04 TLS, equipped with an array of 10 NVIDIA RTX 2080Ti GPUs. 
\subsection{Simulation Settings}
The simulation settings are illustrated in this paragraph. All simulations employ the CosineAnnealingLR as the learning rate adjustment strategy, with a total of 1500 training steps. These hyperparameters are set identically for all optimizers: $\epsilon=1 \mathrm{e}-8$, $\beta_1=0.9$, $\beta_2=0.999$, $\alpha=1 \mathrm{e}-3$, and $weight\_decay =5\mathrm{e}-5$. For the simulations on the loss landscape Fig. \ref{fig.2}(a), the hyperparameters $\kappa$ and $\zeta$ for MIAdam are set to 0.885 and 1400, respectively. The same settings are applied for the simulations on the loss landscape Fig. \ref{fig.2}(e), with the only difference being the learning rate $\alpha=0.005$.
\subsection{Experiment Settings}
\textbf{Image Classification Experiments:}  For MIAdam, we adopt the same hyperparameters as those used for Adam to ensure a consistent comparison basis. Additionally, the hyperparameters $\kappa$ and $\zeta$ are identified with their optimal values through grid search within the range $[0.01,0.02, ..., 0.99]$ and $[1,2, ..., 150]$. After grid search, the optimal values of $\kappa$ and $\zeta$ are 0.98 and 20, respectively. For each optimizer, we calculate the mean and standard deviation of the top-1 test accuracy based on three individual runs while computing the average training time for three runs. For image classification experiments on the CIFAR datasets, the hyperparameter configurations for various optimizers are provided in Table \ref{set:image}.
\begin{table}[h]
   \centering
   \caption{Hyperparameter settings for optimizers of image classification experiments.}
   \small
   \setlength{\tabcolsep}{4pt}
   \begin{tabularx}{0.8\linewidth}{@{} l X @{}}
   \toprule 
   & Epoch $=150$; batch size $=128$; milestone $=[50,75]$ \\
   \hline 
   Adam & $\alpha=1 \mathrm{e}-3, \epsilon=1 \mathrm{e}-8$, $\beta_1=0.9$, $\beta_2=0.999$, $weight\_decay=5\mathrm{e}-5$ \\
   NAdam & $\alpha=1 \mathrm{e}-3, \epsilon=1 \mathrm{e}-8$, $\beta_1=0.9$, $\beta_2=0.999$, $weight\_decay=5\mathrm{e}-5$, $\gamma=1\mathrm{e}-3$\\
   ND-Adam & $\alpha=1 \mathrm{e}-3, \epsilon=1 \mathrm{e}-8$, $\beta_1=0.9$, $\beta_2=0.999$, $weight\_decay=5\mathrm{e}-5$\\
   Adamax & $\alpha=2 \mathrm{e}-3, \epsilon=1 \mathrm{e}-8$, $\beta_1=0.9$, $\beta_2=0.999$, $weight\_decay=5\mathrm{e}-5$\\
   AdaBound & $\alpha=1 \mathrm{e}-3, \epsilon=1 \mathrm{e}-8$, $\beta_1=0.9$, $\beta_2=0.999$, $weight\_decay=5\mathrm{e}-5$\\
   SWATS & $\alpha=1 \mathrm{e}-3, \epsilon=1 \mathrm{e}-8$, $\beta_1=0.9$, $\beta_2=0.999$, $weight\_decay=5\mathrm{e}-5$\\
   Adai & $\alpha=1 \mathrm{e}-3, \epsilon=1 \mathrm{e}-8$, $\beta_1=0.9$, $\beta_2=0.999$, $weight\_decay=5\mathrm{e}-5$\\
   MIAdam & $\alpha=1 \mathrm{e}-3, \epsilon=1 \mathrm{e}-8$, $\beta_1=0.9$, $\beta_2=0.999$, $weight\_decay=5\mathrm{e}-5$, $\kappa=0.98$, $\zeta=20$\\
   \bottomrule
   \end{tabularx}\label{set:image}
   \end{table}

For the experiments conducted on the ImageNet-1k dataset, all settings remain consistent with those used for the CIFAR dataset experiments, except for the total number of training epochs, which is set to 90.  

\textbf{Text Classification Experiments:} We fine-tune the BERT and RoBERTa models using a learning rate of 0.0001 and a batch size of 128 for 30 epochs. For the two optimizers used in training, Adam and MIAdam, both employ milestones as the learning rate adjustment strategy, with the same hyperparameter settings as follows: $\alpha=1 \mathrm{e}-5$, $\epsilon=1 \mathrm{e}-8$, $\beta_1=0.9$, $\beta_2=0.999$, $\alpha=1 \mathrm{e}-3$, and $weight\_decay =5\mathrm{e}-5$. Additionally, for MIAdam, the extra hyperparameters $\kappa$ and $\zeta$ are set to 0.98 and 20, respectively.

\textbf{Experiments on Datasets Injected with Label Noises:}  The extra hyperparameter of MIAdam $\zeta$ is set to 40. Apart from the injection of label noises on the CIFAR-10 dataset, the experimental settings are identical to those of the image classification experiments. 

\section{Supplementary Experiments and Simulations}
\subsection{Simulations}
\begin{figure}[h]
    \centering
        \includegraphics[scale=0.28]{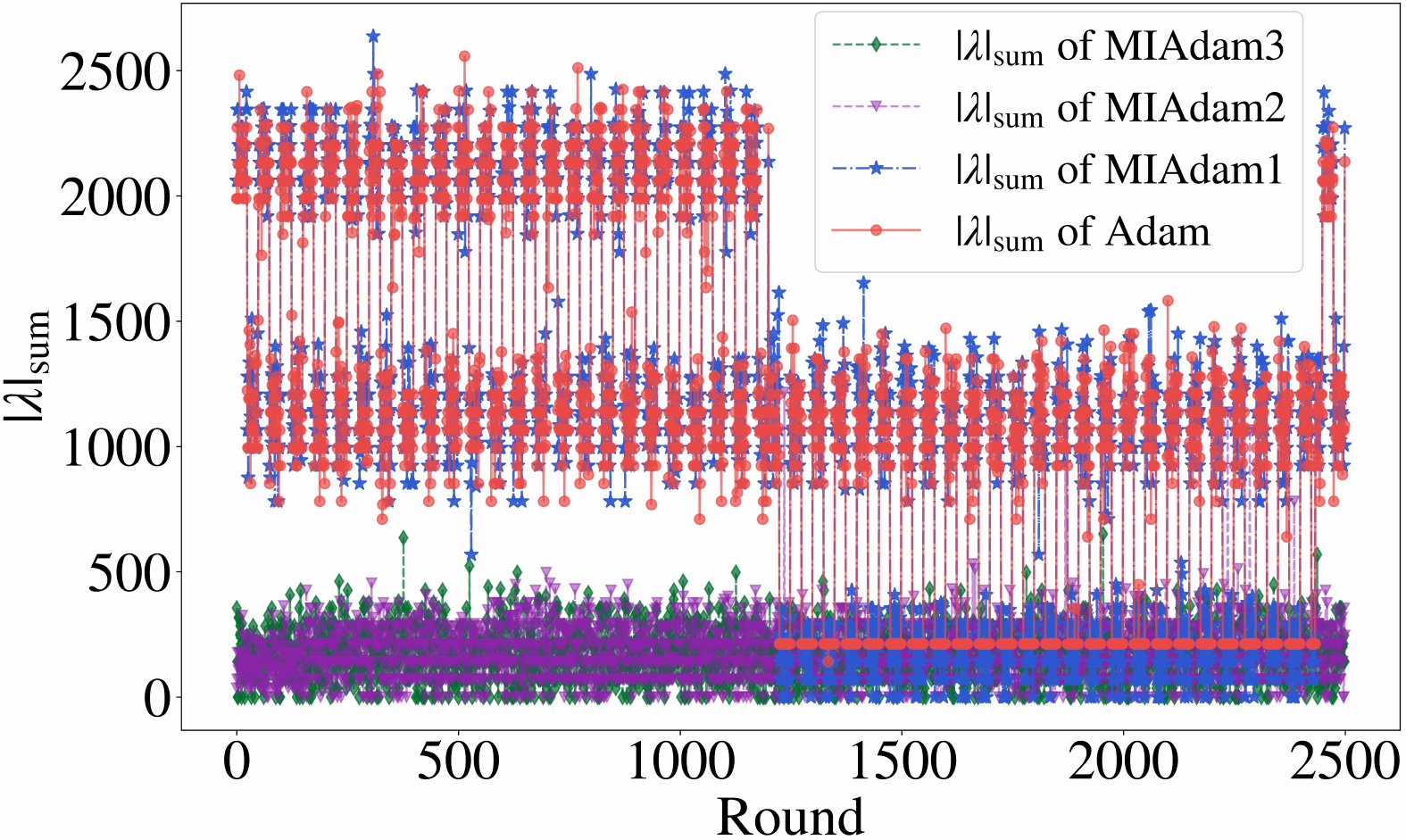} 
      \caption{Comparisons of the sum of the absolute values of the eigenvalues of the Hessian matrix at the convergence points of Adam, MIAdam1,  MIAdam2, and MIAdam3 for 2500 rounds of simulation in the loss landscape shown in Fig. \ref{fig.2}(e).}
          \label{fig:2500run}
\end{figure}

On the second loss landscape in Fig. \ref{fig.2}(e), different starting points influence the trajectory of the optimizer. Therefore, to make the simulation results more convincing, we uniformly select 2,500 starting coordinate points on this loss landscape and conduct 2,500 rounds of simulations in the region where $\theta_1 \in [-2,3]$ and $\theta_2 \in[-2,3]$. Furthermore, we calculate the sum of the absolute values of the eigenvalues of the Hessian matrix of the different optimizers for the final convergence points at different starting coordinate points and compare them. As can be seen in Fig. \ref{fig:2500run}, the sum of the absolute values of the eigenvalues at the location of the final convergence on the second loss landscape is generally smaller for the MIAdam2 and MIAdam3 compared to Adam, which suggests that the multiple integral term is indeed helpful in finding flat minima.
\subsection{Experiments}
\begin{figure}[t]\centering
    \subfigure[]
    {
        \includegraphics[width=0.4055\textwidth]{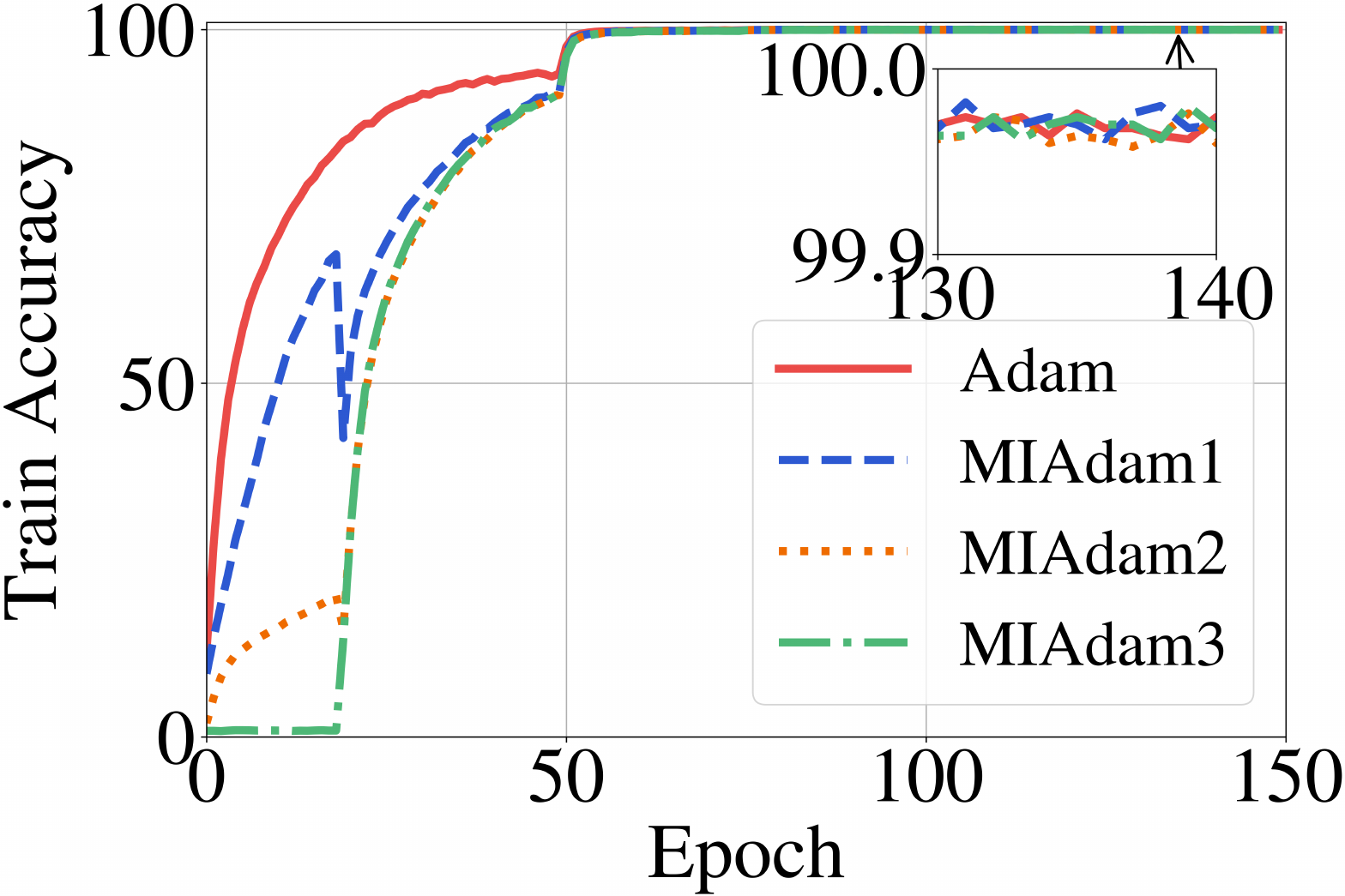}
    }
    \subfigure[]
    {
        \includegraphics[width=0.40\textwidth]{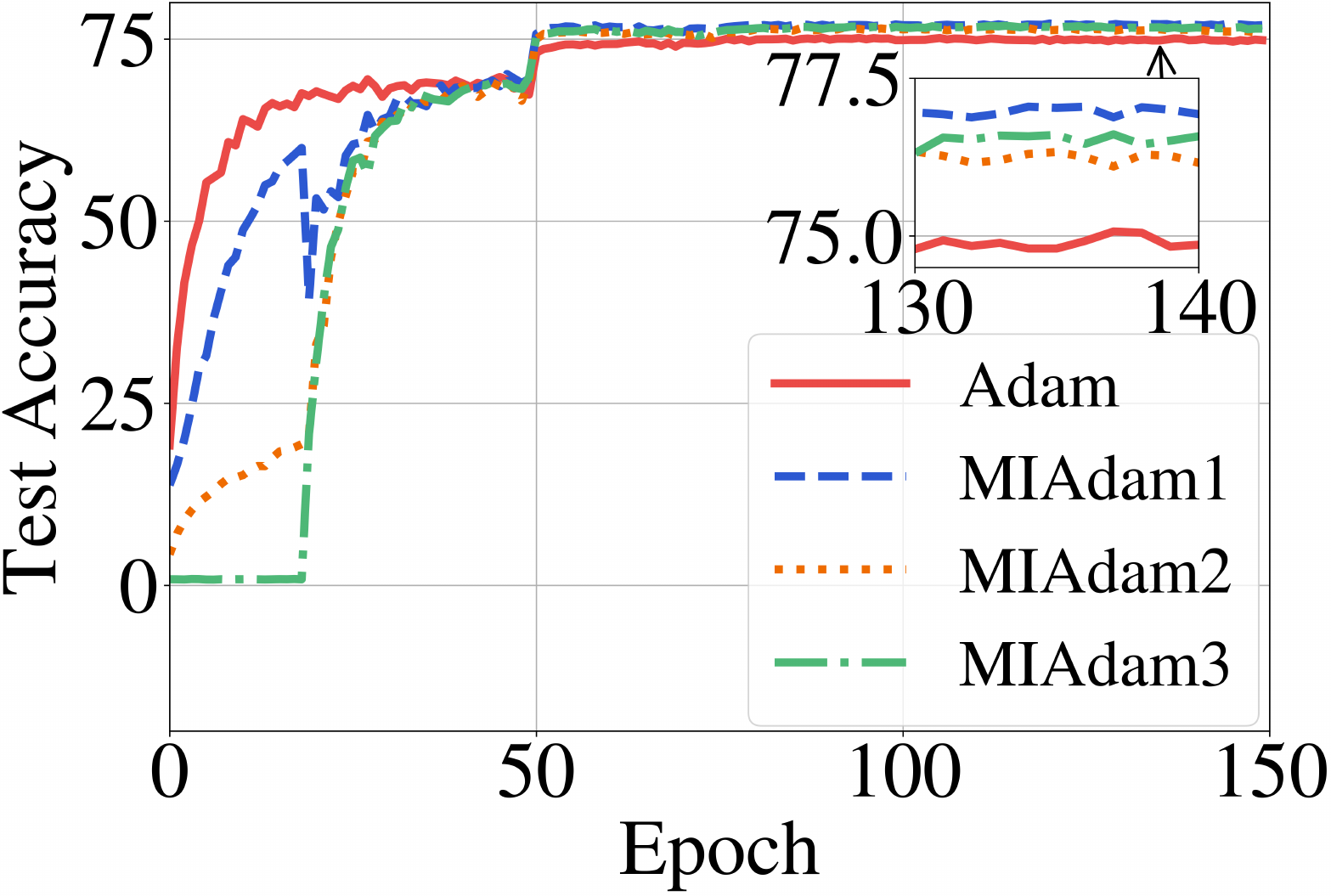}
    }
    \caption{Training and testing comparations of Adam, MIAdam1, MIAdam2, and MIAdam3 on CIFAR-100 using DenseNet121. }
    \label{fig.5}
\end{figure}

Fig. \ref{fig.5} provides training and testing comparisons of Adam, MIAdam1, MIAdam2, and MIAdam3 on CIFAR-100 using DenseNet121. Fig. \ref{fig.5} shows that the introduction of the multiple integral term leads to an enhancement in generalization rather than in training accuracy. It is worth noting that before switching optimizers, training with the multiple integral term leads to slower convergence or even non-convergence, as the optimizer searches for the flat minima on the loss landscape. However, when considering the entire training process, the epochs taken for convergence to a steady state by Adam and MIAdam are similar.
\section{Generalization Proof}\label{proof:gen}
\subsection{Proof of Theorem \ref{thm:generalizaion}}
\begin{proof}
    Without loss of generality, we set the hyperparameter $\kappa=1$ and consider the adaptive learning rate at the end of the derivation to make it convenient to analyze the effect of multiple integral term on generalization. Consequently, the parameter update formula for MIAdam1 without adaptive learning rate and switching optimizers is simplified as follows:
\begin{equation}
\left\{
  \begin{aligned}
&\bm{m}_{t}=\beta_1 \bm{m}_{t-1}+(1-\beta_1)\bm{g}_{t},\\
&\overline{\bm{m}}_{t}^{(1)}= \sum^t_{j=0}\bm{m}_{j},\\
&\bm{\theta}_{t+1}=\bm{\theta}_{t}-{\alpha}^2\overline{\bm{m}}_{t}^{(1)}.
\end{aligned}
\right.
\end{equation} 
 Then, we write the deformed motion equation as follows:
\begin{equation}
\left\{
  \begin{aligned}
&\bm{\mathcal{v}}_t=(1-\delta \Delta t)\bm{\mathcal{v}}_{t-1}+\frac{\bm{\mathcal{f}}}{\chi}\Delta t,\\
&\overline{\bm{\mathcal{v}}}_{t}= \sum^t_{j=0}\bm{\mathcal{v}}_{j},\\
&\bm{\mathcal{p}}_{t+1}=\bm{\mathcal{p}}_{t}+\overline{\bm{\mathcal{v}}}_{t}{(\Delta t)}^2,\\
\end{aligned}
\right.
\end{equation} 
where $\bm{\mathcal{v}}_t =-\bm{m}_{t}$, $\bm{\mathcal{f}} = \bm{g}_{t}$, $\Delta t $ is equivalent to $ \alpha$, $\delta=(1-\beta_1)/\Delta t$, and  $\chi=\Delta t/(1-\beta_1)$.
 When the learning rate $\alpha$ is sufficiently small, the differential form of the motion equation is derived as
\begin{equation}
 \label{eq2}
  \begin{aligned}
\chi\frac{\text{d}^3\bm{\mathcal{p}}(t)}{\text{d}\tilde{t}^3}=\delta \chi\frac{\text{d}^2\bm{\mathcal{p}}(t)}{\text{d}\tilde{t}^2}+\bm{\mathcal{f}},
\end{aligned}
\end{equation} 
where $\text{d}\tilde{t}=\Delta t$. As $\bm{\mathcal{f}}$ corresponds to the stochastic gradient term, we obtain 
\begin{equation}
 \label{eq2}
  \begin{aligned}
\overline{\chi}\text{d}\dot{\bm{\mathcal{p}}}=\delta \overline{\chi}\text{d}\bm{\mathcal{p}}+\frac{\partial L(\bm{\mathcal{p}})}{\partial\bm{\mathcal{p}}}\text{d}\tilde{t}+2[\psi(\bm{\mathcal{p}})]^{\frac{1}{2}}\text{d}W_{\tilde{t}},
\end{aligned}
\end{equation}
where $\overline{\chi}=\chi/\tilde{t}$; $\text{d} W_{\tilde{t}} \sim\mathscr{N}(0, I\text{d}\tilde{t})$; $I$ presents the identity matrix; $\psi(\bm{\mathcal{p}})=\alpha C(\bm{\mathcal{p}})/2$ is the diffusion matrix in the dynamics; $C(\bm{\mathcal{p}})$ denotes the gradient noise covariance matrix. Furthermore, its Fokker-Planck equation in the phase space (the $\dot{\bm{\mathcal{p}}}$-$\bm{\mathcal{p}}$ space) is written as follows:
\begin{equation}
 \label{eq2}
  \begin{aligned}
\frac{\partial P(\bm{\mathcal{p}},\bm{\mathcal{v}},\tilde{t})}{\partial \tilde{t}}&=-\nabla_{\bm{\mathcal{p}}}\cdot[\bm{r}P(\bm{\mathcal{p}},\bm{\mathcal{v}},\tilde{t})]\\
&+\nabla_{\bm{r}} \cdot[\delta \bm{\mathcal{v}}+\overline{\chi}\nabla_{\bm{\mathcal{p}}} L(\bm{\mathcal{p}})]\\
&+\nabla_{\bm{\mathcal{v}}}\overline{\chi}^{-2}\psi\cdot \nabla_{\bm{\mathcal{v}}}P(\bm{\mathcal{p}},\bm{\mathcal{v}},\tilde{t})\\
&=-\nabla \cdot \bm{\varphi}(\bm{\mathcal{p}}, \tilde{t}),
\end{aligned}
\end{equation}
where $\bm{r}=\dot{\bm{\mathcal{p}}}$;  $\bm{\varphi}$ represents the probability flux density; $\nabla \cdot$ denotes the divergence operator. Under Assumption \ref{asu:2}, the probability distribution in the valley around $\bm{a}$ is derived as
\begin{equation}
 \label{eq2}
  \begin{aligned}
&P(\bm{\mathcal{p}},\bm{\mathcal{v}}, \tilde{t})\\
=&\int_{\bm{\mathcal{p}} \in V_{\bm{a}}}\text{exp}[-\frac{\delta\overline{\chi}}{2}(\bm{\mathcal{p}}-{\bm{a}})^{\top}(\psi_{\bm{a}}^{-\frac{1}{2}}H_{\bm{a}} \psi_{\bm{a}}^{-\frac{1}{2}})(\bm{\mathcal{p}}-{\bm{a}})] \text{d}V\\
=&P({\bm{a}})\frac{(2\pi \delta \overline{\chi})^2}{\operatorname{det}(\psi_{\bm{a}}^{-1}H_{\bm{a}})^{\frac{1}{2}}}.
\end{aligned}
\end{equation}
According to \cite{kalinay2012phase}, in the case of finite inertia, we transform the phase-space equation into the position-space Smoluchowski-like form with the effective diffusion correction:
\begin{equation}
 \label{eq2}
  \begin{aligned}
\hat{\psi}_i(\bm{\mathcal{p}}_i)=\psi_{i}(\bm{\mathcal{p}}_i) \left(1-\sqrt{1-\frac{4H_i(\bm{\mathcal{p}}_i)}{\delta^2 \overline{\chi}}}\right) \left( \frac{2H_i (\bm{\mathcal{p}}_i)}{\delta^2\overline{\chi}}\right).
\end{aligned}
\end{equation}
Suppose that $\bm{\varphi}$ is fixed on an escape path from sharp minimum $\bm{a}$ to flat minimum $\bm{b}$ through saddle point $\bm{u}$, we derive $\bm{\varphi}$ from ~\cite{xie2022adaptive} as
\begin{equation}
 \label{eq2}
  \begin{aligned}
\bm{\varphi}=&\frac{\text{exp}\left(\frac{L({\bm{a}})-L(\bm{u})}{T_{\bm{a}}}P({\bm{a}})\right)}{\int^{\bm{b}}_{\bm{a}} \hat{\psi}^{-1}\text{exp}(\frac{L(\bm{\mathcal{p}})-L({{\bm{\varrho}}})}{T})\text{d}\bm{\mathcal{p}}}\\
      =&\frac{\text{exp}\left(\frac{L({\bm{a}})-L(\bm{u})}{T_{\bm{a}}}P({\bm{a}})\right)}{\hat{\psi}_{\bm{u}}^{-1} \text{exp}\left( \frac{L(\bm{u})-L(\bm{\varrho})}{T_{\bm{u}}} \sqrt{\frac{2\pi T_{\bm{u}}}{|H_{\bm{u}}|}}\right)},
\end{aligned}
\end{equation}
where $T=\psi/(\delta\overline{\chi})$ and denotes the temperature parameter in the stationary distribution. Based on the formula of probability current and flux, we obtain the flux escaping through saddle point ${\bm{u}}$:
\begin{equation}
 \label{eq2}
  \begin{aligned}
&\int_{\bm{s}_{\bm{u}}} \bm{\varphi} \cdot \text{d} \bm{s}\\
=& \varphi_i \int_{\bm{s}_{\bm{u}}} \text{exp} \left[ -\frac{\delta \overline{\chi}}{2}(\bm{\mathcal{p}} - \bm{u})^{\top}[\psi_{\bm{u}}^{-\frac{1}{2}}H_{\bm{u}} \psi_{\bm{u}}^{-\frac{1}{2}}]^{\bot \bm{e}}(\bm{\mathcal{p}} -\bm{u} ) \right] \text{d} \bm{s}\\
=& \frac{\text{exp}\left( \frac{L(\bm{a})-L(\bm{\varrho})}{T_{{\bm{a}}{\bm{e}}}}\right)P({\bm{a}})\frac{(2\pi \delta \chi)^{\frac{n-1}{2}}}{(\prod_{i\neq e}(\psi_{\bm{u}_i}^{-1}H_{\bm{u}_i}))^{\frac{1}{2}}}}{\hat{\psi}_{\bm{ue}}^{-1} \text{exp}\left( \frac{L(\bm{u})-L(\bm{\varrho})}{T_{\bm{ue}}} \sqrt{\frac{2\pi T_{\bm{ue}}}{|H_{\bm{ue}}|}}\right)},
\end{aligned}
\end{equation}
where superscript $^{\bot \bm{e}}$ indicates the directions perpendicular to the escape direction $\bm{e}$. Considering the adaptive learning rate, we have $\psi_{\text{MIAdam1}}= \alpha(\sqrt{[H]^{+})/\mathcal{b}})/2$ according to the case of Adam in ~\cite{xie2022adaptive}, where superscript $^+$ presents the transformation that $H = U \operatorname{diag}(H_1, H_2, \cdots , H_{n-1}, H_{n})U^{\top}$ and the $i$-th column vector of $U$  is the eigenvector corresponding to $H_i$.
Finally, we obtain the mean escape time $\phi$ of MIAdam1:
\begin{equation}
 \label{eq2}
\begin{aligned}
\phi_{\text{MIAdam1}}  =& \frac{P\left(\bm{\mathcal{p}} \in V_{\bm{a}}\right)}{\int_{S_b} \bm{\varphi} \cdot \text{d} \bm{s}} \\
 =& \pi\left[\sqrt{1+\frac{4 \alpha \sqrt{\mathcal{b}\left|H_{\bm{u} \bm{e}}\right|}}{\tilde{t}(1-\beta_1)}}+1\right] \frac{\left|\operatorname{det}\left(H_{\bm{a}}^{-1} H_{\bm{u}}\right)\right|^{\frac{1}{4}}}{\left|H_{\bm{u} \bm{e}}\right|}\exp \left[\frac{2 \sqrt{\mathcal{b}} \Delta L}{\tilde{t}\alpha}\left(\frac{\varrho}{\sqrt{H_{{\bm{a}} \bm{e}}}}+\frac{(1-\varrho)}{\sqrt{\left|H_{\bm{u} \bm{e}}\right|}}\right)\right] .
\end{aligned}
\end{equation}
The proof is thus completed.
\end{proof}
\section{Convergence Proof}\label{poof:conv}
\begin{definition}\label{def: conv}
 For a convex set $F$, a function $f: \mathbb{R}^d \rightarrow \mathbb{R}$ is convex, if for all $\bm{x}, \bm{y} \in \mathbb{R}^{d}$ satisfy
\begin{equation}
f\left(\bm{x}\right) \leq f\left(\bm{x}\right)+\nabla f\left(\bm{x}\right)\left(\bm{y}-\bm{x}\right),
\end{equation}
and for all $\lambda \in[0,1]$ satisfy
\begin{equation}
f\left(\lambda \bm{x}+\left(1-\lambda\right) \bm{y}\right) \leq \lambda f\left(\bm{x}\right)+\left(1-\lambda\right) f\left(\bm{y}\right) .
\end{equation}
\end{definition}
\begin{definition}\label{def: RT}
For a convex function $f_t$, we give the $R(\hat{t})$ to determine the convergence. Its expression is shown as follows:
\begin{equation}
  \begin{aligned}
    R\left(\hat{t}\right)=\sum_{t=1}^{\hat{t}} f_t\left(\bm{\theta}_{t}\right)-\min _{\bm{\theta}} \sum_{t=1}^{\hat{t}} f_t\left(\bm{\theta}\right).
  \end{aligned}
\end{equation}
When $\lim _{\hat{t} \rightarrow \infty} R\left({\hat{t}}\right)/{\hat{t}} \neq 0$, we consider the algorithm is not convergent.
\end{definition}
\subsection{Proof of Theorem \ref{thm:convergence}} 
\begin{proof}
 According to the Definition \ref{def: conv} and the Definition \ref{def: RT}, we have\\
\begin{equation}\label{eq2}
  \begin{aligned}
    R\left({\hat{t}}\right)=&\sum_{t=1}^{\hat{t}} [f_t\left(\bm{\theta}_{t}\right)-f_t\left(\bm{\theta}^*\right)]\\
        \leq& \sum_{i=1}^{d}\sum_{t=1}^{\hat{t}} g_{t,i}\left(\theta_{t,i}-\theta_{,i}^{*}\right).
  \end{aligned}
\end{equation}
From the updating rules of MIAdam, we get
\begin{equation}
  \begin{aligned}
      \theta_{t+1,i}=&\theta_{t,i}-\alpha \frac{\hat{m}^{(1)}_{t,i}}{\sqrt{\hat{v}_{t,i}}}\\
                    =&\theta_{t,i}-\alpha \frac{1}{1-\beta_1^t}\frac{\overline{m}^{(1)}_{t,i}}{\sqrt{\hat{v}_{t,i}}}\\
                    =&\theta_{t,i}-\alpha\frac{1}{1-\beta_1^t}\frac{\sum^t_{r=1}\kappa_1^{t-r}m_{r}}{\sqrt{\hat{v}_{t,i}}}.
  \end{aligned}
\end{equation}
Then we obtain
\begin{equation}
  \begin{aligned}\label{eq:23}
      &g_{t,i}\left(\theta_{t,i}-\theta_{,i}^{*}\right)\\=&\underbrace{\frac{(1-\beta_1^t)\sqrt{\hat{v}_{t,i}}}{2\alpha\left(1-\beta_{1,t}\right)}\left(\left(\theta_{t,i}-\theta_{,i}^{*}\right)^2-\left(\theta_{t+1,i}-\theta_{,i}^{*}\right)^{2}\right)}_{\{1\}}\\
+&\underbrace{\frac{\beta_{1,t}m_{t-1,i}\left(\theta_{,i}^{*}-\theta_{t,i}\right)}{\left(1-\beta_{1,t}\right)}}_{\{2\}}+\underbrace{\frac{\kappa_1\overline{m}^{(1)}_{t-1,i}\left(\theta_{,i}^{*}-\theta_{t,i}\right)}{\left(1-\beta_{1,t}\right)}}_{\{3\}}\\   +&\underbrace{\frac{\alpha(1-\beta_1^t)}{2\left(1-\beta_{1,t}\right)}  \frac{\hat{m}_{t,i}^2}{\sqrt{\hat{v}_{t,i}}}}_{\{4\}}.\\
  \end{aligned}
\end{equation}
For the first two terms \{1\} and \{2\} in Eq. (\ref{eq:23}), we derive the following two inequalities based on \cite{kingma2014adam}:
\begin{equation}
  \begin{aligned}
\sum_{i=1}^{d}&\sum_{t=1}^{\hat{t}}\frac{(1-\beta_1^t)\sqrt{\hat{v}_{t,i}}}{2\alpha\left(1-\beta_{1,t}\right)}\left(\left(\theta_{t,i}-\theta_{,i}^{*}\right)^2-\left(\theta_{t+1,i}-\theta_{,i}^{*}\right)^{2}\right)\\ 
\leq& \frac{\mathsf{d}^2}{2\alpha(1-\beta_1)}\sum^d_{i=1}\sqrt{{\hat{t}}\hat{v}_{{\hat{t}},i}},
  \end{aligned}
\end{equation}
and
\begin{equation}
  \begin{aligned}
\sum_{i=1}^{d}&\sum_{t=1}^{\hat{t}}\frac{\beta_{1,t}m_{t-1,i}\left(\theta_{,i}^{*}-\theta_{t,i}\right)}{\left(1-\beta_{1,t}\right)}\\
\leq& \frac{\alpha(1+\beta_1)\mathsf{g}_\infty}{(1-\beta_1)\sqrt{1-\beta_2}(1-\gamma)^2}\sum^d_{i=1}\lVert g_{1:{\hat{t}},i} \rVert_2
+&\sum_{i=1}^{d}\frac{\mathsf{d}^2_{\infty}\mathsf{g}_{\infty}\sqrt{1-\beta_2}}{2\alpha\beta_1(1-\lambda)^2}.
  \end{aligned}
\end{equation}
Since the gradient is assumed to be bounded, the $m_{t,i }$ should also be bounded in Adam and satisfies $m_{t,i } \leq \mathsf{g}$.
For the term \{3\}, we have
\begin{equation}
\begin{aligned}
      &\sum^{\hat{t}}_{t=2}\frac{\kappa_1\overline{m}^{(1)}_{t-1,i}\left(\theta_{,i}^{*}-\theta_{t,i}\right)}{\left(1-\beta_{1,t}\right)}\\
=&\sum^{\hat{t}}_{t=2}\frac{\left(\theta_{,i}^{*}-\theta_{t,i}\right)}{\left(1-\beta_{1,t}\right)}\sum^{t-1}_{r=1}\kappa_1^{t-r}m_{r}\\
\leq&\sum^{\hat{t}}_{t=2}\frac{\left(\theta_{,i}^{*}-\theta_{t,i}\right)}{\left(1-\beta_{1,t}\right)}\left(\frac{\kappa_1}{1-\kappa_1}-\frac{\kappa^t}{1-\kappa_1}\right)t\mathsf{g},
  \end{aligned}
\end{equation}
which is a divergent series. Therefore, we further deduce the following limit:
\begin{equation}
\begin{aligned}
      \lim_{{\hat{t}}\rightarrow \infty}\frac{\sum_{i=1}^{d}\sum^{\hat{t}}_{t=2}\frac{  \kappa_1\overline{m}^{(1)}_{t-1,i}\left(\theta_{,i}^{*}-\theta_{t,i}\right)}{\left(1-\beta_{1,t}\right)}}{{\hat{t}}}\rightarrow \infty.
\end{aligned}
\end{equation}
For the term \{4\}, we have
\begin{equation}
\begin{aligned}
&\sum_{i=1}^{d}\sum^{\hat{t}}_{t=1}\frac{\alpha(1-\beta_1^t)}{2\left(1-\beta_{1,t}\right)}  \frac{\hat{m}_{t,i}^2}{\sqrt{\hat{v}_{t,i}}}\\
 =&\sum_{i=1}^{d}\sum^{\hat{t}}_{t=1}\frac{\alpha(1-\beta_1^t)}{2\left(1-\beta_{1,t}\right)}  \frac{(\sum^{t-1}_{r=1}\kappa_1^{t-r}m_{r})^2}{\sqrt{\hat{v}_{t,i}}}.
\end{aligned}
\end{equation}
Analogous to the derivation of the term \{3\}, we similarly obtain the following limit:
\begin{equation}
\begin{aligned}
      \lim_{{\hat{t}}\rightarrow \infty}\frac{\sum_{i=1}^{d}\sum^{\hat{t}}_{t=1}\frac{\alpha(1-\beta_1^t)}{2\left(1-\beta_{1,t}\right)}  \frac{\hat{m}_{t,i}^2}{\sqrt{\hat{v}_{t,i}}}}{{\hat{t}}}\rightarrow \infty.
\end{aligned}
\end{equation}

Finally, following the above derivation, we have\begin{equation}
\begin{aligned}
\lim_{{\hat{t}}\rightarrow \infty}\frac{R({\hat{t}})}{{\hat{t}}} \neq 0.
  \end{aligned}
\end{equation}
The proof is thus completed.
\end{proof}

\end{document}